\documentclass[smallcondensed]{svjour3}

\smartqed

\usepackage[T1]{fontenc}
\usepackage{aecompl}
\usepackage{color}
\usepackage{amsmath,bm}
\usepackage{mathptmx}
\usepackage{graphicx}

\usepackage{booktabs}
\usepackage[english]{babel}
\usepackage{caption3}
\usepackage{threeparttable}
\usepackage{multirow}
\usepackage{subfigure}
\usepackage{cite}
\usepackage{lineno}

\usepackage{url}
\makeatletter
\def\UrlAlphabet{%
      \do\a\do\b\do\c\do\d\do\e\do\f\do\g\do\h\do\i\do\j%
      \do\k\do\l\do\m\do\n\do\o\do\p\do\q\do\r\do\s\do\t%
      \do\u\do\v\do\w\do\x\do\y\do\z\do\A\do\B\do\C\do\D%
      \do\E\do\F\do\G\do\H\do\I\do\J\do\K\do\L\do\M\do\N%
      \do\O\do\P\do\Q\do\R\do\S\do\T\do\U\do\V\do\W\do\X%
      \do\Y\do\Z}
\def\UrlDigits{\do\1\do\2\do\3\do\4\do\5\do\6\do\7\do\8\do\9\do\0}
\g@addto@macro{\UrlBreaks}{\UrlOrds}
\g@addto@macro{\UrlBreaks}{\UrlAlphabet}
\g@addto@macro{\UrlBreaks}{\UrlDigits}
\makeatother

\journalname{Neural Computing $\&$ Applications}

\begin{document}

\title{Whale swarm algorithm with the mechanism of identifying and escaping from extreme points for multimodal function optimization}

\author{Bing Zeng \and Xinyu Li \and Liang Gao \and Yuyan Zhang \and Haozhen Dong}

\institute{Liang Gao \at
                State Key Laboratory of Digital Manufacturing Equipment and Technology, School of Mechanical Science and Engineering, Huazhong University of Science and Technology, 1037 Luoyu Road, Wuhan, China. \\
                Tel.: +86-027-87559419 \\
                \email{gaoliang@mail.hust.edu.cn}}

\maketitle

\begin{abstract}
Most real-world optimization problems often come with multiple global optima or local optima. Therefore, increasing niching metaheuristic algorithms, which devote to finding multiple optima in a single run, are developed to solve these multimodal optimization problems. However, there are two difficulties urgently to be solved for most existing niching metaheuristic algorithms: how to set the niching parameter valules for different optimization problems, and how to jump out of the local optima efficiently. These two difficulties limit their practicality largely. Based on Whale Swarm Algorithm (WSA) we proposed previously, this paper presents a new multimodal optimizer named WSA with Iterative Counter (WSA-IC) to address these two difficulties. On the one hand, WSA-IC improves the iteration rule of the original WSA for multimodal optimization, which removes the need of specifying different values of attenuation coefficient for different problems to form multiple subpopulations, without introducing any niching parameter. On the other hand, WSA-IC enables the identification of extreme points during the iterations relying on two new parameters (i.e., stability threshold $T_s$ and fitness threshold $T_f$), to jump out of the located extreme points. Moreover, the convergence of WSA-IC is proved. Finally, the proposed WSA-IC is compared with several niching metaheuristic algorithms on CEC2015 niching benchmark test functions and on five additional high-dimensional multimodal functions. The experimental results demonstrate that WSA-IC statistically outperforms other niching metaheuristic algorithms on most test functions.
\end{abstract}

\section{Introduction}\label{sec:Introduction}

Most of the real-world optimization problems are multimodal \cite{tasgetiren2017iterated,lin2016effective,zhang2017Object,ciancio2016heuristic,csevkli2017multi,yi2016parallel,palmieri2017comparison,Zhang2012Nature}, i.e., their objective functions have multiple global optima or local optima. If applying traditional numerical methods to such problems, we have to try many times for locating a different optimum in each run to pick out the best one, which is time-consuming and labor-intensive. In such a scenario, using metaheuristic algorithms, no matter evolutionary algorithms (EAs) or swarm based algorithms, to solve these problems has become a hot research topic, as they are easy to implement and can get as good as possible solutions. However, many metaheuristic algorithms, such as Genetic Algorithm (GA), Particle Swarm Optimization (PSO), Differential Evolution (DE), and so on, are primarily designed to search for a single global optimum. And it is desirable to locate multiple global optima for engineers to choose the most appropriate one. In addition, some metaheuristic algorithms are easy to fall into the local optima. So, many techniques have been proposed for the metaheuristic algorithms to find as many global optima as possible. These techniques are commonly known as niching methods \cite{li2010niching}, which are committed to promoting and maintaining the formation of multiple stable subpopulations within a single population for locating multiple optima. Some representative niching methods include crowding \cite{de1975analysis}, fitness sharing \cite{goldberg1987genetic}, clustering \cite{yin1993fast}, restricted tournament selection \cite{harik1995finding}, parallelization \cite{bessaou2000island}, speciation \cite{deb1989investigation}, and population topologies \cite{kennedy2002population}, and so on. Several of them are presented below, more references and discussions about niching methods can be found in literature \cite{li2016seeking}.

Crowding was firstly proposed by De Jong \cite{de1975analysis} to preserve genetic diversity, so as to improve the global search ability of the algorithm for locating multiple optima. In crowding method, the offspring with better fitness replaces the most similar individual from a subset (i.e., crowd) of the population. The similarity is generally measured by hamming distance for binary encoding and Euclidean distance for real-valued encoding \cite{thomsen2004multimodal}, which means that the smaller the distance between two individuals is, the more similar they are. The individuals of subset are randomly selected from the population, and the size of subset is a user specified parameter called crowding factor ($CF$) that is often set to 2 or 3. However, low $CF$ values will lead to replacement errors, i.e., the offspring replaces another individual with small similarity, which will reduce the population diversity. To avoid replacement errors, deterministic crowding \cite{mahfoud1992crowding} and probabilistic crowding \cite{mengshoel1999probabilistic} were proposed. Setting $CF$ equal to the population size also proved to be effective \cite{thomsen2004multimodal}.

Goldberg and Richardson \cite{goldberg1987genetic} proposed fitness sharing mechanism, which enables the formation of multiple subpopulations by formulating sharing functions. When using this method, the shared fitness of all the individuals need to be calculated according to Eq.\ref{eq:shared fitness}.

\begin{equation}\label{eq:shared fitness}
f_i^{'} = \frac{{{f_i}}}{{m_i^{'}}}
\end{equation}

\noindent where, $f_i$ and $f_i^{'}$ are the original fitness and shared fitness of individual $i$ respectively; $m_i^{'}$ is the shared value of individual $i$ with other individuals, and is formulated as $m_i^{'} = \sum\limits_{j = 1}^N {sh\left( {{d_{ij}}} \right)}$, where $N$ is the population size, $sh(d_{ij})$ is the sharing function over the individual $i$ and $j$, which is calculated as follows.

\begin{equation}\label{eq:sharing function}
sh\left( {{d_{ij}}} \right) = \left\{ {\begin{array}{*{20}{l}}
{1 - {{\left( {\frac{{{d_{ij}}}}{{{\sigma _{share}}}}} \right)}^\alpha }}&{{\rm{if}}\;{d_{ij}} < {\sigma _{share}},}\\
0&{{\rm{otherwise}}{\rm{.}}}
\end{array}} \right.
\end{equation}

\noindent where, $\alpha$ is a constant, and always set as 1; $d_{ij}$ is the distance between the individual $i$ and $j$; $\sigma _{share}$ is the sharing distance, which is always set as the value of peak radius. However, this method assumes that all the peaks have the equal height and width. Obviously, a prior knowledge of the fitness landscape is required to set the value of $\sigma _{share}$.

\begin{figure}[htbp]
    \setlength{\abovecaptionskip}{0pt}
    \subfigure[two individuals follow the same extreme point]{
        \label{fig:detect-multimodal:subfig:a}
        \begin{minipage}[c]{0.48\textwidth}
            \centering
            \includegraphics[width=1.0\textwidth]{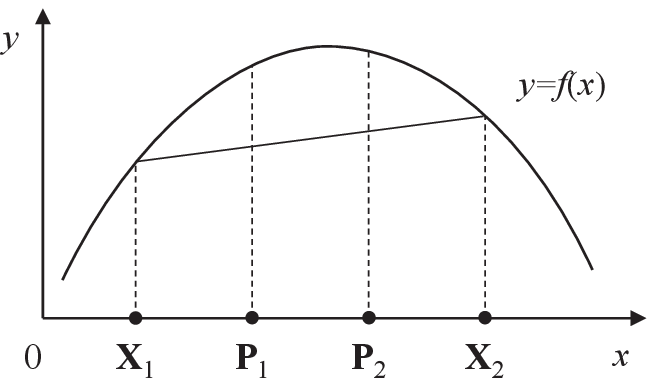}
        \end{minipage}}%
    \subfigure[two individuals follow different extreme points]{
        \label{fig:detect-multimodal:subfig:b}
        \begin{minipage}[c]{0.48\textwidth}
            \centering
            \includegraphics[width=1.0\textwidth]{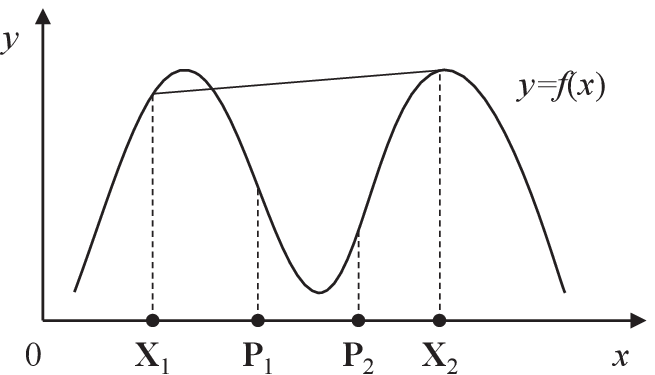}
        \end{minipage}}
    \caption{Sketch maps of the ``detect-multimodal'' mechanism}
    \label{fig:detect-multimodal:subfig}
\end{figure}

Speciation \cite{deb1989investigation} is another popular niching technique, which is used to form parallel subpopulations, i.e., species, according to the similarity between individuals. The similarity is also measured by distance, such as Euclidean distance. This niching technique employs one user-specified parameter called species distance ($\sigma _s$) to divide the population into a set of species. It is assumed that the problem to be solved is a maximization optimization problem. The detailed procedure of forming species in every generation is shown below. The first step is to find out the species seeds that dominate their own species. Firstly, an empty set $\bf{X}_s$ is defined to contain the species seeds. Sorting the individuals in decreasing order of fitness and adding the first individual of population after sorting to the set $\bf{X}_s$. Then, judging the remaining individuals one by one in order, and determining whether they are within the distance of $\sigma _s$/2 from any species seed in $\bf{X}_s$. If no, they are added to $\bf{X}_s$. After all the individuals are traversed, the set $\bf{X}_s$ has collected all the species seeds. Next comes the step of adding the individuals to their corresponding species. For each species seed in $\bf{X}_s$, adding the individuals that are within the distance of $\sigma _s$/2 from it to its species, if an individual has been added to a species, doing nothing. Although speciation method is able to divide the population into multiple subpopulations, it has a major shortcoming. Its parameter, i.e., species distance, is hard to set precisely for different optimization problems. In such case, inspired by the Multinational Evolutionary Algorithms \cite{ursem1999multinational}, Stoean et al. \cite{stoean2007disburdening} proposed ``detect-multimodal'' mechanism to establish species, which removes the need of specifying distance parameter. The ``detect-multimodal'' mechanism utilizes a set of interior points between two individuals to detect whether there is a valley between them in the fitness landscape, so as to determine whether the two individuals track different extreme points. If all the interior points are better than the worse one of these two individuals, they are considered to follow the same extreme point, i.e., locating in the same peak of the fitness landscape, as shown in Fig. \ref{fig:detect-multimodal:subfig:a}, wherein, $f(\mathbf{P}_1)$>$f(\mathbf{X}_1)$ and $f(\mathbf{P}_2)$>$f(\mathbf{X}_1)$. On the contrary, if there exist at least one interior point that is worse than the worse one of these two individuals, at least one valley is considered existing between the two individuals, i.e., they are considered to track different extreme points as shown in Fig. \ref{fig:detect-multimodal:subfig:b}, wherein, $f(\mathbf{P}_1)$<$f(\mathbf{X}_1)$. Those individuals following the same extreme point are added to the same species. Although ``detect-multimodal'' mechanism does not utilize species distance to divide the population into multiple species, it employs another parameter called ``number of gradations'', i.e., number of interior points, which also depends on the problem characteristics.

Thus it can be seen that some niching methods need to set some parameters, which require prior knowledge of the fitness landscape, to divide the population into multiple subpopulations. However, for many real-world optimization problems, the prior knowledge of the fitness landscape is very difficult or almost impossible to obtain \cite{li2010niching}. Therefore, these niching methods are difficult to be used to deal with the real-world optimization problems. In this paper, a new multimodal optimization algorithm called Whale Swarm Algorithm with Iterative Counter (WSA-IC), based on our preliminary work in \cite{Zeng2017}, is proposed. By improving the iteration rule of the original WSA for multimodal optimization, WSA-IC removes the need of specifying parameter values for different problems to form multiple subpopulations, without introducing any niching parameter. In addition, WSA-IC enables the identification of extreme point to jump out of the located extreme points during the iterations.

The remainder of this paper is organized as follows. A brief overview of the multimodal optimization algorithms is presented in section \ref{sec:Related works}. Section \ref{sec:WSA} introduces WSA briefly. A detailed description of the proposed WSA-IC is presented in section \ref{sec:WSA-IC}. The next section presents the experimental results and analysis to evaluate WSA-IC. The last section draws the conclusions and presents the future research.

\section{Related works}\label{sec:Related works}

With increasing niching methods put forward, a large number of multimodal optimization algorithms combining the metaheuristic algorithms with these niching methods have been proposed. In this section, a brief overview of multimodal optimization algorithms is presented. According to whether the prior knowledge of the fitness landscape is needed, these multimodal optimization algorithms are classified into prior knowledge based methods and non-prior knowledge based methods. More references and discussions about multimodal optimization algorithms can be found in literatures \cite{li2016seeking,das2011real}.

\subsection{Prior knowledge based methods}

Species Conserving Genetic Algorithm (SCGA) was proposed by Li et al. \cite{li2002species} via introducing speciation and species conservation techniques into the classical GA. In each iteration, the current population is partitioned into multiple subpopulations (i.e., species) using the speciation technique \cite{deb1989investigation}, before executing the genetic operators. Moreover, after executing the genetic operators, all the species seeds are either conserved to the next generation or replaced by better members of the same species, which can contribute significantly to the preservation of global and local optima that have been found so far. Li showed that the additional overhead of SCGA caused by these two techniques was not higher than that introduced by Genetic Algorithm with Sharing (SGA) \cite{goldberg1987genetic}, and SCGA performs far better than SGA in success rates of locating the global optima.

Li \cite{li2005efficient} proposed Species-based DE (SDE) algorithm to solve multimodal optimization problems via introducing speciation technique. In SDE algorithm, when the number of member individuals of a species is less than a predefined value, the algorithm will randomly generate new individuals within the radius of species seed until the species size reaches the predefined value. Then, the conventional DE algorithm is implemented separately for each identified species. In addition, if the fitness of an offspring is the same as that of its species seed, this offspring will be replaced by a randomly generated new individual. These two mechanisms improved the efficiency of SDE algorithm significantly.

The speciation technique was also introduced into the conventional PSO by Li \cite{li2004adaptively} to solve multimodal optimization problems. In each iteration of Species-based PSO (SPSO), after the population is divided into multiple species and the species seeds are determined, each species seed is assigned to its member individuals as the $lbest$. Then, each individual updates its position according to the iterative equations concerning velocity and position of the $lbest$ PSO. The experimental results showed that SPSO was comparable to or better than SNGA \cite{beasley1993sequential}, SCGA and NichePSO \cite{brits2002niching} over a set of multimodal functions.

Stoean et al. \cite{stoean2007disburdening} proposed Topological Species Conservation (TSC) algorithm, which utilizes the ``detect-multimodal'' mechanism to remove the need of specifying distance parameter when selecting species seeds and forming species. In TSC algorithm, all the individuals that track the same extreme point are in the same species, which corresponds to the real structure of the optimization function. And the species seeds can also be conserved to the next generation. However, TSC algorithm need excessive fitness evaluations in seeds selection procedure, especially when the number of interior points get larger. For improving the computational efficiency of TSC algorithm, i.e., saving the fitness evaluations, Stoean et al. \cite{stoean2010multimodal} proposed Topological Species Conservation Version 2 (TSC2) algorithm. In TSC2 algorithm, the current unclassified individual chooses the seed one by one in ascending order of distance from it to perform the ``detect-multimodal'' procedure until the return value is true or this individual is considered a new seed, because the species dominated by the closer seed is more likely to track the same peak with the current individual. Through this method, TSC2 algorithm saves considerable fitness evaluations. In addition, when the optimization function has a large number of local optima, TSC algorithm might pick out too many seeds from the population that would be conserved to the next generation, significantly reducing the search ability of TSC algorithm. And TSC2 algorithm introduced the maximum number of seeds to guarantee the algorithm's search ability.

Deb and Saha \cite{deb2010finding} firstly converted a single-objective multimodal optimization problem into a bi-objective optimization problem. Multiple global and local optima of the original problem become the members of weak Pareto-optimal set of the transformed problem. One of the objectives of the transformed problem is the objective function of the original problem. With regards to the other objective, the gradient-based approach is firstly employed, which is based on the property that the derivatives of objective function at the minimum points are equal to zero. However, the derivatives of objective function at the maximum and saddle points are also equal to zero, and the objective functions of some optimization problems may be non-differentiable at the minimum points. Then, more pragmatic neighborhood count based approaches are developed for establishing the second objective, which is the number of neighboring solutions that are better than the current solution. During the iterations, the non-dominated ranks of different solutions rely on two parameters, i.e., $\sigma _f$ and $\sigma _x$, which are used to distinguish two optima.

\subsection{Non-prior knowledge based methods}

Thomsen \cite{thomsen2004multimodal} proposed Crowding-based DE (CDE) algorithm by introducing crowding method into the conventional DE for multimodal function optimization. In CDE algorithm, the similarity of two individuals is measured by the Euclidean distance between two individuals. The fitness value of an offspring is only compared with that of the most similar individual in the current population, and the offspring replaces the most similar individual if it has better fitness. This replacement scheme can make the population remain diversity in the search space, which makes a great contribution to the location of multiple optima. Thomsen showed that CDE algorithm performed better than a fitness sharing DE variant over a group of multimodal functions.

The History based topological speciation (HTS) was proposed by Li and Tang \cite{li2015history} to incorporate into the CDE with species conservation technique for multimodal optimization. HTS is a parameter-free speciation method, which captures the landscape topography relying exclusively on search history. As a result, it avoids the additional sampling and function evaluations associated with existing topology based methods. Therefore, HTS is a parameter-free speciation method. The experimental results showed that HTS performed better than existing topology-based methods when the function evaluation budget is limited.

Liang et al. \cite{liang2006comprehensive} proposed Comprehensive Learning Particle Swarm Optimizer (CLPSO) for multimodal function optimization. In CLPSO, all particles' best previous positions can potentially be used to guide a particle's flying, i.e., each dimension of a particle may learn from the corresponding dimension of different particle's best previous position. The velocity updating equation of CLPSO is shown as follows.

\begin{equation}\label{eq:CLPSO velocity updating}
V_i^d = \omega  * V_i^d + c * rand_i^d * \left( {pbest_{{f_i}\left( d \right)}^d - X_i^d} \right)
\end{equation}

\noindent where, $\omega$ is an inertia weight, $c$ is an acceleration constant, $X_i^d$ denotes the $d$-th dimension of particle $i$'s position, $V_i^d$ represents the $d$-th dimension of particle $i$'s velocity. $rand_i^d$ is a random number between 0 and 1 associated with $X_i^d$. For particle $i$, a set $\bm{f}_i$=[$f_i(1)$, $f_i(2)$, $\cdots$, $f_i(d)$, $\cdots$, $f_i(D)$], where $D$ denotes the dimension of fitness function, is built to store the serial numbers of those particles whose best previous positions particle $i$ should learn from at the corresponding dimensions. $pbest_{{f_i}\left( d \right)}^d$ denotes the $d$-th dimension of particle $f_i(d)$'s best previous position. The values of elements in $\bm{f}_i$ depend on the learning probability $P_c$ that can take different values for different particles. For example, generate a random number for assigning $f_i(d)$. If this random number is greater than $P_c^i$, assign $i$ to $f_i(d)$; otherwise, assign the serial number of a particle selected from population through tournament selection procedure to $f_i(d)$. If particle $i$ does not find a better position after a certain number of iterations called the refreshing gap $m$, reassign $\bm{f}_i$ for particle $i$.

Li \cite{li2007multimodal} proposed Fitness-Distance-Ratio based PSO (FERPSO) algorithm, which utilizes FER to avoid specifying any niching parameter, for multimodal function optimization. The FER value with respect to particle $i$ and particle $j$ is shown as follows.

\begin{equation}\label{eq:FER value}
{\rm{FE}}{{\rm{R}}_{\left( {j,i} \right)}} = \alpha  \cdot \frac{{f\left( {{{\overrightarrow {\bf{P}} }_j}} \right) - f\left( {{{\overrightarrow {\bf{P}} }_i}} \right)}}{{\left\| {{{\overrightarrow {\bf{P}} }_j} - {{\overrightarrow {\bf{P}} }_i}} \right\|}}
\end{equation}

\noindent where, ${\overrightarrow {\bf{P}} _i}$ and ${\overrightarrow {\bf{P}} _j}$ are the best previous positions of particle $i$ and particle $j$ respectively; $\alpha$ is a scaling factor and formulated as follows.

\begin{equation}\label{eq:scaling factor}
\alpha  = \frac{{\left\| s \right\|}}{{f\left( {{{\overrightarrow {\bf{P}} }_g}} \right) - f\left( {{{\overrightarrow {\bf{P}} }_w}} \right)}}
\end{equation}

\noindent where, ${\overrightarrow {\bf{P}} _g}$ and ${\overrightarrow {\bf{P}} _w}$ are the best particle and worst particle in current population respectively. ||$s$|| is the size of search space, which is estimated by its diagonal distance $\sqrt {\sum\nolimits_{k = 1}^{Dim} {{{\left( {x_k^u - x_k^l} \right)}^2}} } $ (where $Dim$ denotes the dimension of search space, i.e., the number of variables. $x_k^u$ and $x_k^l$ are the upper and lower bounds of the $k$-th variable $x_k$, respectively). In every iteration, each particle needs to calculate the FER value with respect to it and every other particle to find the neighboring point denoted by ${\overrightarrow {\bf{P}} _n}$, corresponding to the maximal FER value. Then, each particle updates its velocity according to Eq. \ref{eq:FERPSO velocity updating}. Over successive iterations, some subpopulations tracking different peaks will be formed, so as to locate multiple optima.

\begin{equation}\label{eq:FERPSO velocity updating}
{\overrightarrow {\bf{v}} _i} = \chi \left( {{{\overrightarrow {\bf{v}} }_i} + {{\overrightarrow {\bf{R}} }_1}\left[ {0,\;{{{\varphi _{max}}} \mathord{\left/
 {\vphantom {{{\varphi _{max}}} 2}} \right.
 \kern-\nulldelimiterspace} 2}} \right] \otimes \left( {{{\overrightarrow {\bf{p}} }_i} - {{\overrightarrow {\bf{x}} }_i}} \right) + {{\overrightarrow {\bf{R}} }_2}\left[ {0,{{\;{\varphi _{max}}} \mathord{\left/
 {\vphantom {{\;{\varphi _{max}}} 2}} \right.
 \kern-\nulldelimiterspace} 2}} \right] \otimes \left( {{{\overrightarrow {\bf{p}} }_n} - {{\overrightarrow {\bf{x}} }_i}} \right)} \right)
\end{equation}

\noindent where, ${\overrightarrow {\bf{v}} _i}$ and ${\overrightarrow {\bf{x}} _i}$ are the velocity and position of particle $i$ respectively. ${\overrightarrow {\bf{R}} _1}[ 0,\;$ ${{{\varphi _{max}}} \mathord{\left/ {\vphantom {{{\varphi _{max}}} 2}} \right. \kern-\nulldelimiterspace} 2}]$ and ${\overrightarrow {\bf{R}} _2}\left[ {0,{{\;{\varphi _{max}}} \mathord{\left/ {\vphantom {{\;{\varphi _{max}}} 2}} \right. \kern-\nulldelimiterspace} 2}} \right]$ denote two vectors which are comprised of random values generated between 0 and ${{{\varphi _{max}}} \mathord{\left/ {\vphantom {{{\varphi _{max}}} 2}} \right. \kern-\nulldelimiterspace} 2}$. ${\varphi _{max}}$ is a positive constant. And $\chi $ is a constriction coefficient.

The $lbest$ PSO niching algorithms using ring topology, such as $r3pso$, $r2pso$, $r3pso$-$lhc$ and $r2pso$-$lhc$, were also proposed by Li \cite{li2010niching} for multimodal function optimization. These ring topology based PSO niching algorithms also remove the need of specifying any niching parameters. Taking $r3pso$ for example, a particle's neighboring best point ${\overrightarrow {\bf{P}} _n}$, shown in Eq. \ref{eq:FERPSO velocity updating}, is set as the best one among the best previous positions of its two immediate neighbors (i.e., left and right neighbors identified by population indices). Using the ring topology methods, these $lbest$ PSO algorithms are able to form multiple subpopulations over successive iterations. Li showed that the $lbest$ PSO algorithms using ring topology could provide comparable or better performance than SPSO and FERPSO on some test functions.

Qu et al. \cite{qu2012differential} proposed a neighborhood based mutation and integrated it with three niching DE algorithms, i.e., CDE, SDE and sharing DE \cite{thomsen2004multimodal}, for multimodal function optimization. In neighborhood mutation, the subpopulations are formed, relying on the parameter neighborhood size $m$. During the iterations, each individual should calculate the Euclidean distances from other individuals in the population. Then, selecting the former $m$ nearest individuals form a subpopulation for each individual. And the offspring of each individual is generated by using the corresponding DE algorithm within the subpopulation that the individual belongs to. After a certain number of iterations, some subpopulations will track different extreme points of the multimodal function to be optimized. Generally, the parameter $m$ can be set to a value between 1/20 of the population size and 1/5 of the population size.

The locally informed PSO (LIPS) algorithm was proposed by Qu et al. \cite{qu2013distance} for multimodal function optimization. LIPS makes use of the local information (best previous positions of several neighbors) to guide the search of each particle. The velocity updating equation of LIPS is shown as follows.

\begin{equation}\label{eq:LIPS velocity updating}
V_i^d = \omega  * \left( {V_i^d + \varphi  * \left( {P_i^d - X_i^d} \right)} \right)
\end{equation}

\noindent where, $\omega $ is an inertia weight, $X_i^d$ denotes the $d$-th dimension of particle $i$'s position, $V_i^d$ is the $d$-th dimension of particle $i$'s velocity. ${P_i} = \frac{{{{\sum\limits_{j = 1}^{nsize} {\left( {{\varphi _j} \cdot nbes{t_j}} \right)} } \mathord{\left/ {\vphantom {{\sum\limits_{j = 1}^{nsize} {\left( {{\varphi _j} \cdot nbes{t_j}} \right)} } {nsize}}} \right. \kern-\nulldelimiterspace} {nsize}}}}{\varphi }$, $nsize$ is the neighbor size, which is dynamically increased from 2 to 5 during the iterations; ${\varphi _j}$ is a random number generated in [0, 4.1/$nisze$], and $\varphi  = \sum\limits_{j = 1}^{nsize} {{\varphi _j}} $; $nbest_j$ is the best previous position of the $j$-th nearest neighbor to the $i$-th individual's best previous position. With this technique, LIPS algorithm eliminates the requirement for specifying any niching parameters and improves the local search ability. Qu et al. showed that LIPS algorithm outperformed several well-known niching algorithms, containing $r3pso$, $r2pso$, SPSO, FERPSO, SDE and CDE, and so on, over 30 standard benchmark functions not only on success rate but also with regard to accuracy.

Yazdani et al. \cite{Yazdani2014gravitational} proposed Niche Gravitational Search Algorithm (NGSA) based on the laws of gravity and motion. To find multiple solutions in multimodal problems, the main population of NGSA is partitioned into smaller sub-swarms by introducing three strategies: a $K$-nearest neighbors ($K\textrm{-}NN$) strategy, an elitism strategy and modification of active gravitational mass formulation. The key parameter $K$, i.e., the number of neighbors, is adaptively defined as $K\left( t \right) = Round\left( {\left[ {{K_i} - \left( {{K_i} - {K_f}} \right) \cdot \frac{t}{T}} \right]N} \right)$, where t is the current iteration; $T$ denotes the maximal iterations; $N$ represents the population size; $K_i$ and $K_f$ are two constants that determine the number of neighbors at the beginning and the end of the search, always set to 0.08 and 0.16 respectively.

Wang et al. \cite{wang2015mommop} proposed Multiobjective Optimization for Multimodal Optimization Problems (MOMMOP), which transforms a Multimodal Optimization Problem (MMOP) into a Multiobjective Optimization Problem (MOP) with two conflicting objectives. In this way, all the global optima of the original MMOP can become the Pareto optimal solutions of the transformed problem. With MOMMOP, an MMOP is transformed into a MOP as follows.

\begin{equation}\label{eq:MOMMOP}
\left\{ {\begin{array}{*{20}{l}}
{{\rm{minimize}}}&{{f_1}\left( {\overrightarrow x } \right) = {x_1} + \frac{{\left| {f\left( {\overrightarrow x } \right) - BestOFV} \right|}}{{\left| {WorstOFV - BestOFV} \right|}} \cdot \left( {{U_1} - {L_1}} \right) \cdot \eta }\\
{{\rm{minimize}}}&{{f_2}\left( {\overrightarrow x } \right) = 1 - {x_1} + \frac{{\left| {f\left( {\overrightarrow x } \right) - BestOFV} \right|}}{{\left| {WorstOFV - BestOFV} \right|}} \cdot \left( {{U_1} - {L_1}} \right) \cdot \eta }
\end{array}} \right.
\end{equation}

\noindent where, $\overrightarrow x  = \left( {{x_1},\;{x_2},\; \cdots ,\;\;{x_i},\; \cdots ,\;{x_D}} \right)$ is a solution, ${x_i}$($i \in \left\{ {1,\;2,\; \cdots ,\;D} \right\}$) is the $i$-th variable, and $D$ denotes the number of variables. ${f_1}\left( {\overrightarrow x } \right)$ and ${f_2}\left( {\overrightarrow x } \right)$ are the two conflicting objectives of the transformed problem. $f\left( {\overrightarrow x } \right)$ is the objective function value of $\overrightarrow x$ with respect to the original problem. $BestOFV$ and $WorstOFV$ denote the best and worst objective function values during the evolution, respectively. $U_1$ and $L_1$ are the upper and lower bounds of the first variable, respectively. $\eta$ is the scaling factor, which gradually increases during the evolution. Because some optima may have the same values in certain variables, for the sake of locating multiple global optima, each variable is used to construct a bi-objective optimization problem similar to Eq. \ref{eq:MOMMOP}. If a solution ${\overrightarrow x _u}$ Pareto dominates another solution ${\overrightarrow x _v}$ on all the $D$ bi-objective optimization problems, ${\overrightarrow x _u}$ is considered to dominate ${\overrightarrow x _v}$. What's more, to make the population more evenly distributed, another comparison criterion is proposed. That is a solution ${\overrightarrow x _u}$ dominates another solution ${\overrightarrow x _v}$ if

\begin{equation}\label{eq:domination criteria}
f\left( {{{\overrightarrow x }_u}} \right)\;{\rm{is}}\;{\rm{better\;than }}f\left( {{{\overrightarrow x }_v}} \right)\; \wedge \;distance\left( {normalization\left( {{{\overrightarrow x }_u},\;{{\overrightarrow x }_v}} \right)} \right) < 0.01
\end{equation}

\noindent where, $f\left( {{{\overrightarrow x }_u}} \right)$ and $f\left( {{{\overrightarrow x }_v}} \right)$ are the objective function values of ${\overrightarrow x _u}$ and ${\overrightarrow x _v}$, respectively, with respect to the original problem. $distance\left( {normalization\left( {{{\overrightarrow x }_u},\;{{\overrightarrow x }_v}} \right)} \right)$ denotes the Euclidean distance between the normalized ${\overrightarrow x _u}$ and ${\overrightarrow x _v}$ (i.e., $x_{u,i}$=($x_{u,i}$ $-$ $L_i$)/($U_i$ $-$ $L_i$), $x_{v,i}$=($x_{v,i}$ $-$ $L_i$)/($U_i$ $-$ $L_i$), where $i \in $\{1, $\cdots$,$D$\}). If $distance$({ \emph{normalization}({${{\overrightarrow x }_u}$, ${{\overrightarrow x }_v}$})})<0.01, ${\overrightarrow x _u}$ and ${\overrightarrow x _v}$ is considered to be quite similar to each other.

\subsection{Our motivations}

Based on the above overview, we can find that lots of multimodal optimization algorithms need to set some niching parameters, which require prior knowledge of the fitness landscape. However, this is very difficult or impossible for many real-world optimization problems. What's more, few existing multimodal optimization algorithms can effectively identify and get rid of the located extreme points during the iterations. Since they have no mechanism to determine whether a subpopulation has already located the extreme point of a peak, before the end of running. Therefore, lots of function evaluations will be wasted, when an extreme point has been located early. And it also restricts the global search ability of the algorithm if a subpopulation all the time tracks an extreme point located early.

Based on the above analysis, our main motivations in this paper are summarized as follows.

\begin{enumerate}

\item[1)] Improve the iteration rule of the original WSA to remove the need of specifying different values of attenuation coefficient $\eta$ for different problems to form multiple subpopulations, without adding any niching parameters.
\item[2)] Enable the identification of extreme point and jumping out of the located extreme points during the iterations, relying on two new parameters named stability threshold $T_s$ and fitness threshold $T_f$, so as to eliminate the unnecessary function evaluations and improve the global search ability.

\end{enumerate}

\section{Whale swarm algorithm}\label{sec:WSA}

Inspired by the whales' behavior of communicating with each other via ultrasound for hunting, we proposed WSA for function optimization \cite{Zeng2017}. As shown in our previous work \cite{Zeng2017}, WSA performs well on maintaining population diversity and has strong local search ability, which contribute significantly to locating the global optima with high accuracy. WSA updates the position of a whale $\bf{X}$ under the guidance of its ``better and nearest'' whale $\bf{Y}$, according to the following equation.

\begin{equation}\label{eq:WSA}
x_i^{t{\rm{ + 1}}} = x_i^t + {\rm{rand}}\left( {0,\;{\rho _0} \cdot {e^{ - \eta  \cdot {d_{{\bf{X}},\;{\bf{Y}}}}}}} \right) * \left( {y_i^t - x_i^t} \right)
\end{equation}

\noindent where, $x_i^t$ and $x_i^{t{\rm{ + 1}}}$ denote the $i$-th element of $\bf{X}$'s position at $t$ and $t$+1 iterations respectively, and $y_i^t$ represents the $i$-th element of $\bf{Y}$'s position at $t$ iteration. ${{\rho _0}}$ is the intensity of ultrasound source, which can be set to 2 for almost all the cases. $e$ denotes the natural constant. $\eta$ is the attenuation coefficient. And $d_{\bf{X},\bf{Y}}$ is the Euclidean distance between $\bf{X}$ and $\bf{Y}$. ${\rm{rand}}\left( {0,\;{\rho _0} \cdot {e^{ - \eta  \cdot {d_{{\bf{X}},\;{\bf{Y}}}}}}} \right)$ denotes a random value generated between 0 and ${\rho _0} \cdot {e^{ - \eta  \cdot {d_{{\bf{X}},\;{\bf{Y}}}}}}$ uniformly. According to Eq. \ref{eq:WSA}, a whale would move positively and randomly under the guidance of its ``better and nearest'' whale which is close to it, and move negatively and randomly under the guidance of that whale which is quite far away from it.

\begin{figure}[htbp]
    \setlength{\abovecaptionskip}{0pt}
    \includegraphics[width=0.65\textwidth]{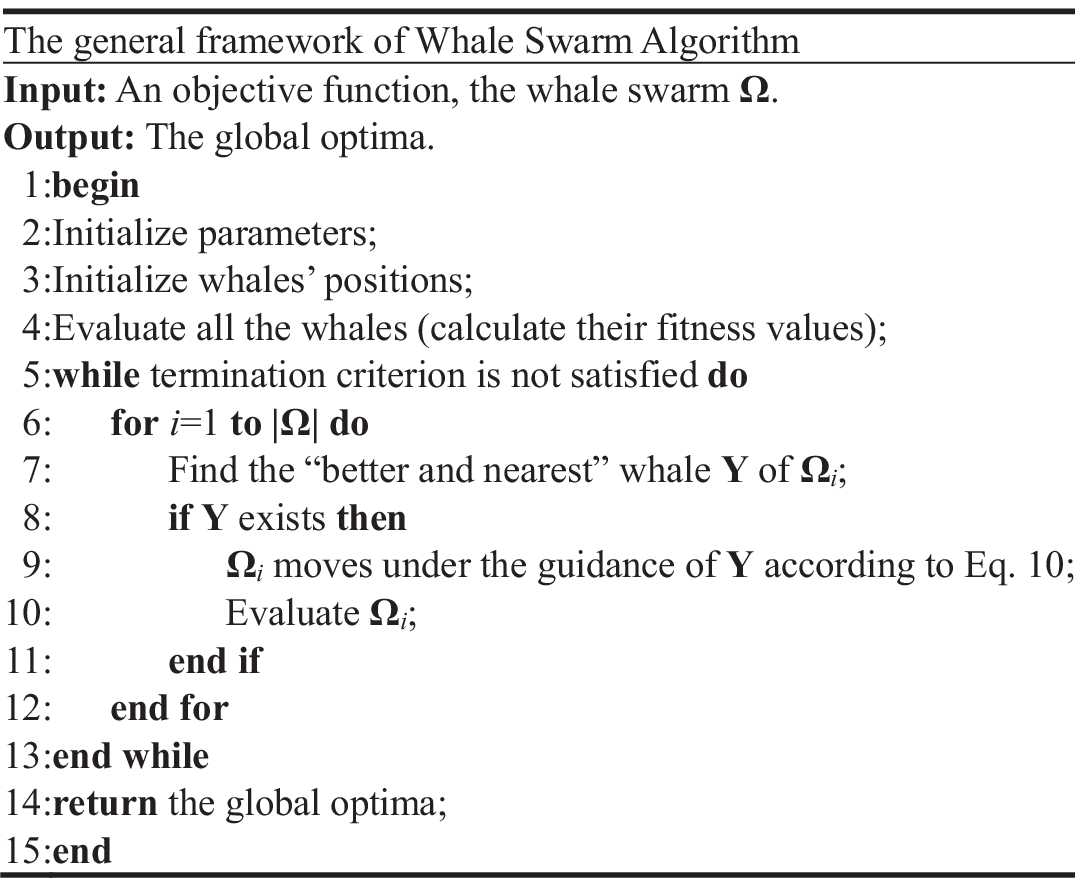}
    \caption{The general framework of WSA}
    \label{fig:The general framework of WSA}
\end{figure}

\begin{figure}[htbp]
    \setlength{\abovecaptionskip}{0pt}
    \includegraphics[width=0.55\textwidth]{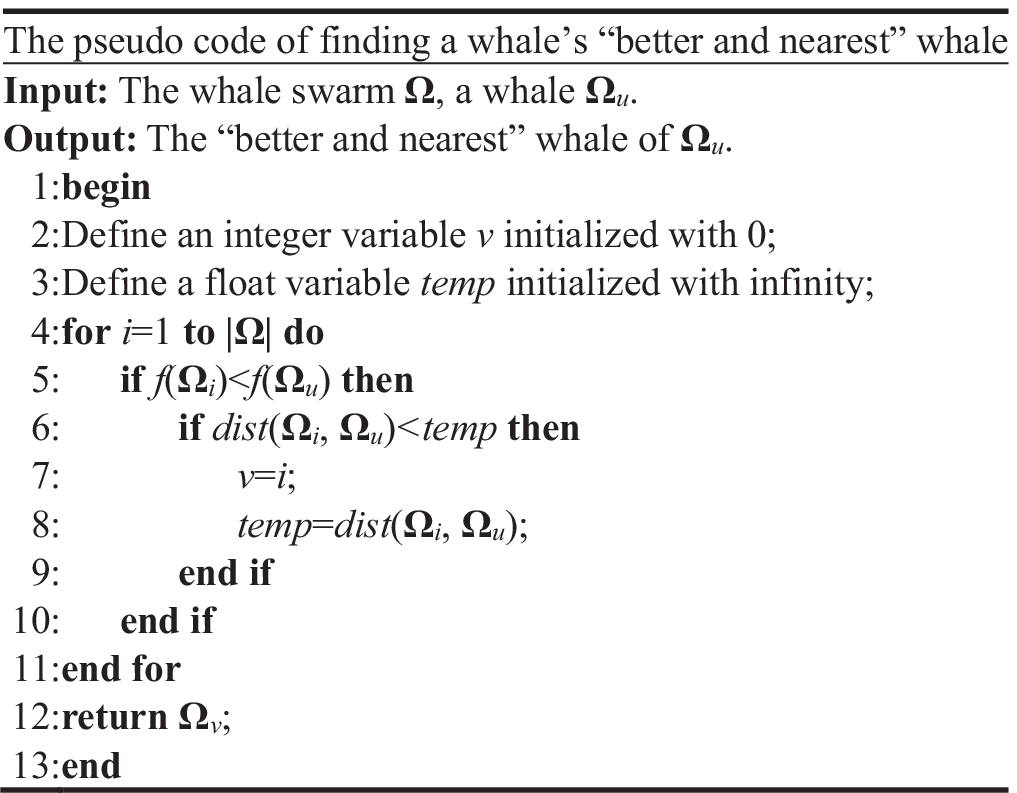}
    \caption{The pseudo code of finding a whale's ``better and nearest'' whale}
    \label{fig:Finding the better and nearest whale}
\end{figure}

The general framework of WSA is shown in Fig. \ref{fig:The general framework of WSA}, where |${\bm{\Omega }}$| in line 6 denotes the number of members in ${\bm{\Omega }}$, namely the swarm size, and ${\bm{\Omega }}_i$ in line 7 is the $i$-th whale in ${\bm{\Omega }}$. From Fig. \ref{fig:The general framework of WSA}, it can be seen that WSA has a fairly simple structure. In every iteration, before moving, each whale needs to find its ``better and nearest'' whale as shown in Fig. \ref{fig:Finding the better and nearest whale}, where $f({\bm{\Omega }}_i)$ in line 6 is the fitness value of whale ${\bm{\Omega }}_i$.

\section{The proposed algorithm (WSA-IC)}\label{sec:WSA-IC}

Firstly, the improvements of WSA for multimodal function optimization are presented in this section. Then, the implementation of WSA-IC is described in sufficient detail. Next, the parameters setting of WSA-IC is discussed. Finally, the convergence analysis of WSA-IC is given. It is assumed that the problems to be solved by the algorithms are minimization problems. Let the fitness functions be the same as the objective functions.

\subsection{The improvements of WSA}\label{sec:improvements}

\begin{enumerate}

\item[1)] The improvement on iteration rule of WSA

\end{enumerate}

\begin{figure}[htbp]
    \setlength{\abovecaptionskip}{0pt}
    \includegraphics[width=0.4\textwidth]{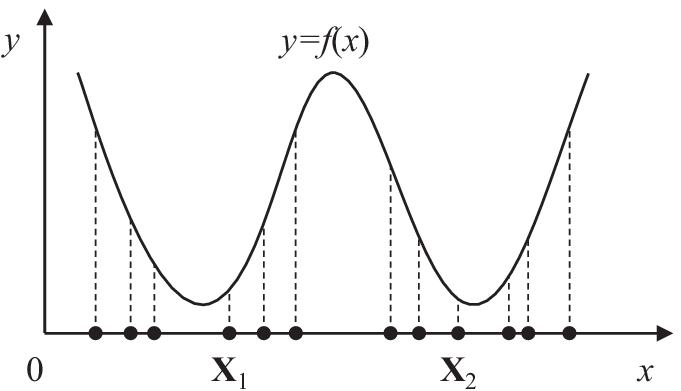}
    \caption{A sketch map of a whale and its ``better and nearest'' whale tracking different extreme points}
    \label{fig:Two whales track different optima}
\end{figure}

Although the original WSA performs well in forming multiple parallel subpopulations and maintaining the population diversity, it needs to specify different values of attenuation coefficient $\eta$ for different problems, which reduces the practicality of WSA. Thus, we improve the iteration rule of WSA to remove the need of specifying different values of attenuation coefficient $\eta$ for different problems, on the premise of ensuring the formation of multiple subpopulations and the ability of local exploitation. Firstly, we assume that the intensity of ultrasound does not attenuate in water, i.e., $\eta$=0, which means that each whale can correctly understand the message sent out by any other whale in the search area. Therefore, a whale will move positively and randomly under the guidance of its ``better and nearest'' whale, regardless of whether that whale is close to it or far away from it. So, when a whale and its ``better and nearest'' whale track different extreme points, the whale may move far away from the extreme point tracked by it due to the guidance of its ``better and nearest'' whale that follows another extreme point, which will weaken WSA's ability of local exploitation. Taking a one-dimensional function optimization problem for example, as shown in Fig. \ref{fig:Two whales track different optima}, the whale ${\bf{X}}_1$ is near to an extreme point, while its ``better and nearest'' whale ${\bf{X}}_2$ is near to another extreme point. In this case, ${\bf{X}}_1$ may move to a worse point or even go to another peak under the guidance of ${\bf{X}}_2$, which will impede the location of the extreme point tracked by ${\bf{X}}_1$ previously. Obviously, this situation is not conducive to locating multiple global optima for WSA.

To solve the above problem effectively, we improved the rule of updating location for each whale as follows. Firstly, generating a copy ${{\bf{X}}^{'}}$ of a whale $\bf{X}$. Then, ${{\bf{X}}^{'}}$ moves under the guidance of $\bf{X}$'s ``better and nearest'' whale $\bf{Y}$ according to Eq. \ref{eq:WSA}. If the position of ${{\bf{X}}^{'}}$ after movement is better than that of $\bf{X}$ (i.e., the fitness value of ${{\bf{X}}^{'}}$ after movement is less than that of $\bf{X}$), $\bf{X}$ will move to ${{\bf{X}}^{'}}$; otherwise, $\bf{X}$ will remain unchanged. In a word, if a whale finds a better position by Eq. \ref{eq:WSA} in an iteration, it will move to the better position; otherwise, it will remain quiescent in its current position, which is similar to the elitism strategy in EAs. So, when it comes to the case shown in Fig. \ref{fig:Two whales track different optima}, the probability of whale ${\bf{X}}_1$ moving away from the extreme point tracked by it will be reduced very much, because it is difficult for whale ${\bf{X}}_1$ to find a better position by Eq. \ref{eq:WSA} under the guidance of its ``better and nearest'' whale ${\bf{X}}_2$. In other words, the whale ${\bf{X}}_1$ may stay at its current position with high probability to guide the movement of other whales. When there exists at least one whale that follows the same extreme point as ${\bf{X}}_1$ and is better than ${\bf{X}}_1$ in the meantime, ${\bf{X}}_1$ will converge to the extreme point under the guidance of the nearest one among those better whales, in next iteration. Therefore, this improvement will contribute significantly to forming multiple subpopulations and enhancing the ability of local exploitation for the improved WSA, which are very conducive to locating multiple global optima, despite $\eta$=0. What's more, this improvement does not introduce any niching parameters.

\begin{enumerate}

\item[2)] Identifying and escaping from the located extreme points during the iterations

\end{enumerate}

In the field of multimodal optimization, identifying the located extreme points effectively and jumping out of these extreme points for saving unnecessary function evaluations during the iterations are very important for metaheuristic algorithms to locate the global optimum/optima. Although the improved WSA mentioned above can ensure the formation of multiple subpopulations and the ability of local exploitation, it cannot yet identify the located extreme points and escape from these extreme points during the iterations. In such case, we propose two new parameters, i.e., stability threshold $T_s$ and fitness threshold $T_f$, which aims to help each whale identify the located optima and jump out of these optima during the iterations, so as to save unnecessary function evaluations and improve the global search ability. $T_s$ is a predefined number of iterations utilized to judge whether a whale has reached steady state, and reaching steady state means that this whale has located the extreme point tracked by it. And $T_f$ is a predefined value utilized to judge whether a solution is a current global optimum. If a whale does not find a better position after successive $T_s$ iterations, it is considered to have reached steady state and located an extreme point. If the difference between its fitness value and $f_{gbest}$ (the fitness value of the best one among the current global optima) is less than $T_f$, the whale's position is considered a current global optimum; otherwise, the whale's position is considered a local optimum. If the whale's position is a current global optimum, this optimum will be stored. Then, the whale that has reached steady state is randomly reinitialized in the search area to jump out of the located extreme point. To judge whether a whale has reached steady state, each whale keeps an iterative counter $c$ to record the number of successive iterations during which it has not found a better position. So, in this paper, the improved WSA is called WSA with Iterative Counter (WSA-IC).

\subsection{The detailed procedure of WSA-IC}\label{sec:detailed procedure}

\begin{figure}[htbp]
    \setlength{\abovecaptionskip}{0pt}
    \includegraphics[width=0.65\textwidth]{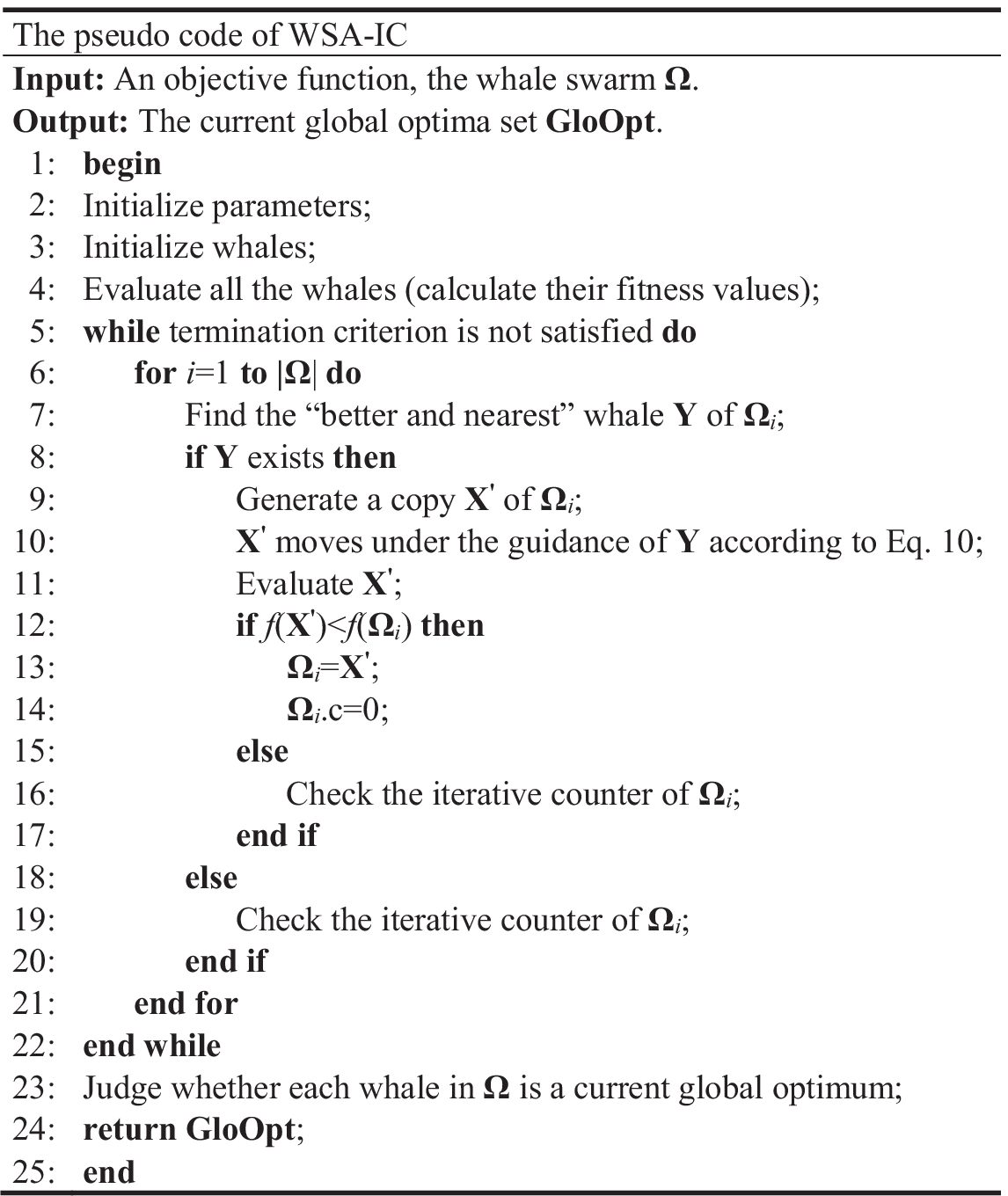}
    \caption{The pseudo code of WSA-IC}
    \label{fig:WSA-IC}
\end{figure}

Fig. \ref{fig:WSA-IC} presents the pseudo code of WSA-IC. For WSA-IC, it is worth noting that the initialization of a whale contains two operations: initializing the whale's position randomly and assigning 0 to its iterative counter. The improvement on iteration rule of WSA described in section \ref{sec:improvements} can be seen from Fig. \ref{fig:WSA-IC}. If a whale's ``better and nearest'' whale exists (line 8 in Fig. \ref{fig:WSA-IC}), a copy of this whale is generated firstly (line 9 in Fig. \ref{fig:WSA-IC}). Then, the copy moves under the guidance of the ``better and nearest'' whale according to Eq. \ref{eq:WSA} (line 10 in Fig. \ref{fig:WSA-IC}). If the position of this copy after movement is better than that of the original whale (line 12 in Fig. \ref{fig:WSA-IC}), the copy replaces the original whale (line 13 in Fig. \ref{fig:WSA-IC}).

\begin{figure}[htbp]
    \setlength{\abovecaptionskip}{0pt}
    \includegraphics[width=0.55\textwidth]{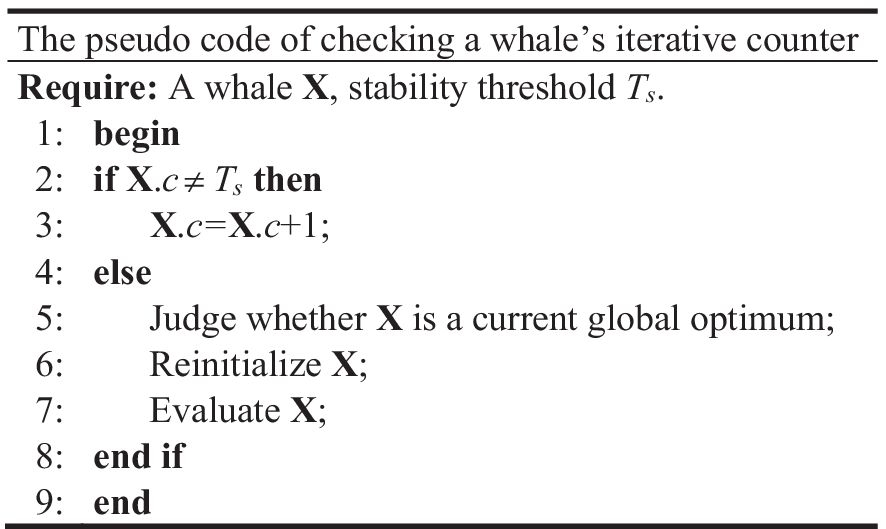}
    \caption{The pseudo code of checking a whale's iterative counter}
    \label{fig:Checking the iterative counter}
\end{figure}

The detail of identifying and escaping from the located extreme points during the iterations for WSA-IC is shown below. If a whale finds a better position (lines $9-13$ in Fig. \ref{fig:WSA-IC}) in an iteration, assigning 0 to its iterative counter $c$ (line 14 in Fig. \ref{fig:WSA-IC}); otherwise, the whale should check its iterative counter (lines $15-17$ and $18-20$ in Fig. \ref{fig:WSA-IC}). The detailed procedure of checking a whale's iterative counter is demonstrated in Fig. \ref{fig:Checking the iterative counter}. As we can see from Fig. \ref{fig:Checking the iterative counter}, firstly determine whether the whale's iterative counter $c$ has reached stability threshold $T_s$. If the whale's iterative counter $c$ is less than $T_s$ (line 2 in Fig. \ref{fig:Checking the iterative counter}), its $c$ increases by 1 (line 3 in Fig. \ref{fig:Checking the iterative counter}); otherwise, the whale is considered to have reached steady state and located an extreme point. If the whale has reached steady state, it should determine whether the located extreme point is a current global optimum (line 5 in Fig. \ref{fig:Checking the iterative counter}). If it is a current global optimum, this extreme point will be stored. Then, the whale that has reached steady state is randomly reinitialized (line 6 in Fig. \ref{fig:Checking the iterative counter}), for jumping out of the located extreme point to find the global optima. It can be seen that, with the parameter stability threshold $T_s$, the proposed WSA-IC can jump out of the located extreme points without hindering local search.

\begin{figure}[htbp]
    \setlength{\abovecaptionskip}{0pt}
    \includegraphics[width=0.75\textwidth]{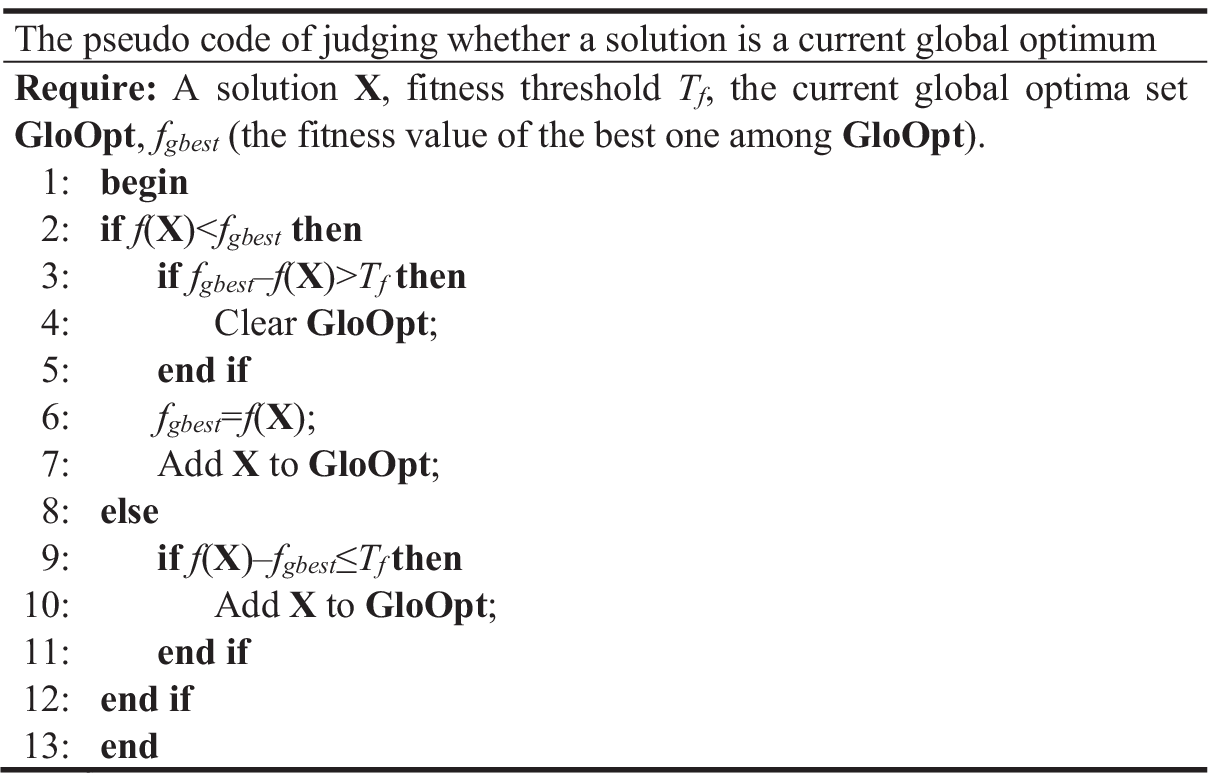}
    \caption{The pseudo code of judging whether a solution is a current global optimum}
    \label{fig:Judging whether a solution is a current global optimum}
\end{figure}

The detailed procedure of judging whether a solution is a current global optimum is demonstrated in Fig. \ref{fig:Judging whether a solution is a current global optimum}. Firstly, judge whether the fitness value of the solution is less than $f_{gbest}$ (the fitness value of the best one among the current global optima set $\bf{GloOpt}$). If the fitness value of this solution is less than $f_{gbest}$ (line 2 in Fig. \ref{fig:Judging whether a solution is a current global optimum}), this solution must be the current global optimum. Before updating $f_{gbest}$ (line 6 in Fig. \ref{fig:Judging whether a solution is a current global optimum}) and storing the new current global optimum (line 7 in Fig. \ref{fig:Judging whether a solution is a current global optimum}), judge whether the optima located before in $\bf{GloOpt}$ are still the current global optima. If the difference between $f_{gbest}$ and the whale's fitness is greater than $T_f$ (line 3 in Fig. \ref{fig:Judging whether a solution is a current global optimum}), all the elements of $\bf{GloOpt}$ are not the current global optima, so $\bf{GloOpt}$ needs to be cleared (line 4 in Fig. \ref{fig:Judging whether a solution is a current global optimum}). If the fitness value of this solution is greater than $f_{gbest}$ (line 8 in Fig. \ref{fig:Judging whether a solution is a current global optimum}), judge whether this solution is a current global optimum. If the difference between the fitness value of this solution and $f_{gbest}$ is not greater than $T_f$ (line 9 in Fig. \ref{fig:Judging whether a solution is a current global optimum}), this solution is considered a current global optimum, so it is added to $\bf{GloOpt}$ (line 10 in Fig. \ref{fig:Judging whether a solution is a current global optimum}).

Until the end of iterations, though some whales' iterative counters do not reach $T_s$, they may have already located the current global optima. Therefore, conducting the step in Fig. \ref{fig:Judging whether a solution is a current global optimum} for each whale in the last generation (line 23 in Fig. \ref{fig:WSA-IC}) is necessary.

\subsection{Parameters setting of WSA-IC}\label{sec:Parameters setting}

\begin{figure}[htbp]
    \setlength{\abovecaptionskip}{0pt}
    \includegraphics[width=0.55\textwidth]{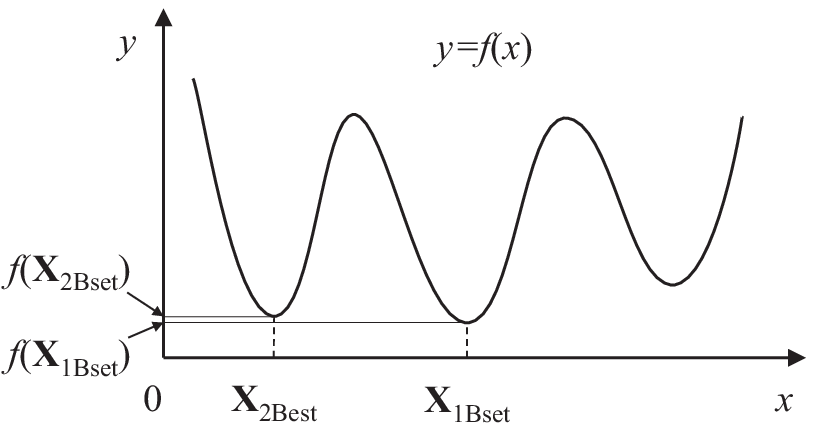}
    \caption{A function with at least one local optimum}
    \label{fig:A function with at least one local optimum}
\end{figure}

As we can see from the detailed steps above, WSA-IC contains four algorithm dependent parameters, i.e., intensity of ultrasound source $\rho _0$, attenuation coefficient $\eta$, stability threshold $T_s$ and fitness threshold $T_f$. $\rho _0$ and $\eta$ are two constants, and are always set to 2 and 0 respectively. $T_f$ should be set to a comparatively small value that is between 0 and the difference between the global second best fitness and the global best fitness, if the problem to be solved has at least one local optimum as shown in the example of an one-dimensional function in Fig. \ref{fig:A function with at least one local optimum}. The ${\bf{X}}_\mathrm{1Best}$ and ${\bf{X}}_\mathrm{2Best}$ in Fig. \ref{fig:A function with at least one local optimum} denote the global optimum and the global second best solution respectively, and the difference between their objective function values is quite small. For the function to be optimized in Fig. \ref{fig:A function with at least one local optimum}, $T_f$ should be set to a very small value that between 0 and $f({\bf{X}}_\mathrm{2Best})-f({\bf{X}}_\mathrm{1Best})$. For almost all the problems, especially those problems without prior knowledge of their fitness landscape, $T_f$ can be set to ${\rm{1}}{\rm{.0}} \times {\rm{1}}{{\rm{0}}^{{\rm{ - 8}}}}$. And for those benchmark test functions whose global optima are given, $T_f$ can be set to the value of the predefined fitness error (i.e., level of accuracy) that is utilized to judge whether a solution is a real global optimum. The value of $T_s$ may vary with the problem to be solved. According to a large number of experimental results, it is reasonable to set $T_s$=100$n$, where $n$ is the function dimension.

\subsection{Convergence analysis of WSA-IC}

It can be seen from section \ref{sec:detailed procedure} that if a whale's iterative counter $c$ increases to $T_s$, the whale is considered to have reached steady state, i.e., it has converged. So, the convergence analysis of WSA-IC depends on the convergence proof of position update rules of WSA-IC. Based on Fig. \ref{fig:WSA-IC} and Eq. \ref{eq:WSA}, the position update equation of WSA-IC can be expressed as follows.

\begin{equation}\label{eq:WSA-IC}
x_i^{t + 1} = \left\{ {\begin{array}{*{20}{l}}
{Ax_i^t + By_i^t{\kern 1pt} }&{f(Ax_i^t + By_i^t) < f(x_i^t),}\\
{x_i^t}&{f(Ax_i^t + By_i^t) \ge f(x_i^t).}
\end{array}} \right.
\end{equation}

\noindent where, $A$ = 1$-$rand(0, 2), $B$ = rand(0, 2). It follows that E($A$) = 0, E($B$) = 1 and D($A$) $=$ D($B$) $=$ $¨C$E($AB$) $=$ 1/3.

To prove the convergence of Eq. \ref{eq:WSA-IC} just needs to prove the convergence of expectation and variance of $x_i^{t + 1}$. The expectation of $x_i^{t + 1}$ is shown as follows.

\begin{equation}\label{eq:expectation}
{\rm{E}}\left( {x_i^{t + 1}} \right) = {\rm{E}}\left( {Ax_i^t + By_i^t} \right)
\end{equation}

Because the distribution of $B$ is unrelated to $x_i^t$ and $y_i^t$, $y_i^t$ can be treated as a constant. And Eq. \ref{eq:expectation} can be rewritten as follows.

\begin{equation}
{\rm{E}}\left( {x_i^{t + 1}} \right) = {\rm{E}}\left( A \right){\rm{E}}\left( {x_i^t} \right) + {\rm{E}}\left( B \right)y_i^t
\end{equation}

\begin{equation}
\frac{1}{{{\rm{E}}\left( B \right)}}{\rm{E}}\left( {x_i^{t + 1}} \right) - \frac{{{\rm{E}}\left( A \right)}}{{{\rm{E}}\left( B \right)}}{\rm{E}}\left( {x_i^t} \right) = y_i^t
\end{equation}

The eigenvalue $\lambda$ of ${\rm{E}}\left( {x_i^{t + 1}} \right)$ satisfies the following characteristic equation.

\begin{equation}\label{eq:characteristic equation}
\frac{1}{{{\rm{E}}\left( B \right)}}{\rm{\lambda }} - \frac{{{\rm{E}}\left( A \right)}}{{{\rm{E}}\left( B \right)}} = 0
\end{equation}

The sufficient and necessary condition for the convergence of ${\rm{E}}\left( {x_i^{t + 1}} \right)$ is that the eigenvalue $\lambda$ is less than 1. It can be seen from Eq. \ref{eq:characteristic equation} that $\lambda$ = E($A$) = 0. Therefore, we can conclude that ${\rm{E}}\left( {x_i^{t + 1}} \right)$ will converge during the iterations.

The variance of $x_i^{t + 1}$ is shown as follows.

\begin{equation}\label{eq:variance}
\begin{array}{l}
{\rm{D}}\left( {x_i^{t + 1}} \right) = {\rm{E}}{\left( {x_i^{t + 1}} \right)^2} - {{\rm{E}}^2}\left( {x_i^{t + 1}} \right) = {\rm{E}}{\left( {Ax_i^t + By_i^t} \right)^2} - {{\rm{E}}^2}\left( {Ax_i^t + By_i^t} \right)\\
 = {\rm{E}}\left( {{A^2}} \right){\rm{E}}{\left( {x_i^t} \right)^2} - {{\rm{E}}^2}\left( A \right){{\rm{E}}^2}\left( {x_i^t} \right) + 2{\rm{E}}\left( {AB} \right){\rm{E}}\left( {x_i^t} \right)y_i^t\\
 - 2{\rm{E}}\left( A \right){\rm{E}}\left( B \right){\rm{E}}\left( {x_i^t} \right)y_i^t + \left( {{\rm{E}}\left( {{B^2}} \right) - {{\rm{E}}^2}\left( B \right)} \right){\left( {y_i^t} \right)^2}
\end{array}
\end{equation}

Eq. \ref{eq:variance} can be transformed as follows.

\begin{equation}\label{eq:variance transform}
\begin{array}{l}
{\rm{D}}\left( {x_i^{t + 1}} \right) - {\rm{E}}\left( {{A^2}} \right){\rm{D}}\left( {x_i^t} \right) = {\rm{D}}\left( A \right){{\rm{E}}^2}\left( {x_i^t} \right) + \\
2{\rm{E}}\left( {AB} \right){\rm{E}}\left( {x_i^t} \right)y_i^t - 2{\rm{E}}\left( A \right){\rm{E}}\left( B \right){\rm{E}}\left( {x_i^t} \right)y_i^t + {\rm{D}}\left( B \right){\left( {y_i^t} \right)^2}\\
 = {\rm{D}}\left( A \right)\left( {{{\rm{E}}^2}\left( {x_i^t} \right) - 2{\rm{E}}\left( {x_i^t} \right)y_i^t + {{\left( {y_i^t} \right)}^2}} \right){\kern 1pt}
\end{array}
\end{equation}

From Eq. \ref{eq:variance transform}, it follows that the eigenvalue $\lambda$ of ${\rm{D}}\left( {x_i^{t + 1}} \right)$ is equal to E($A^2$). So ${\rm{D}}\left( {x_i^{t + 1}} \right)$ will converge during the iterations because E($A^2$) = 1/3 that is less than 1. Therefore, we can expect that during the iterations of WSA-IC, the whales will converge to an appropriate solution under the guidance of their ``better and nearest'' whales.

\section{Experimental results and analysis}\label{sec:Experimental results and analysis}

The proposed WSA-IC and other comparison algorithms are all implemented with C++ programming language by Microsoft visual studio 2015 and executed on the PC with 3.2 GHz and 3.6 GHz Intel core i5-3470 processor, 4 GB RAM and 64-bit Microsoft windows 10 operating system. The source code of the proposed WSA-IC can be download from the website \url{https://drive.google.com/file/d/1W5uUvmdYjKYoC1QsHd5HQkSyZD2Hf0de/view?us
p=sharing}. The five niching metaheuristic comparison algorithms are listed as follows.

\begin{enumerate}

\item[1)] LIPS \cite{qu2013distance}: the locally informed PSO.
\item[2)] NGSA \cite{Yazdani2014gravitational}: the niche GSA.
\item[3)] NSDE \cite{qu2012differential}: the neighborhood based speciation DE.
\item[4)] NCDE \cite{qu2012differential}: the neighborhood based crowding DE.
\item[5)] FERPSO \cite{li2007multimodal}: the Fitness-Euclidean distance ratio PSO.

\end{enumerate}

Apart from the above niching metaheuristic algorithms, WSA-IC is also compared with WSA \cite{Zeng2017}. It is worth noting that the different evolutionary rules of different algorithms will result in different computational complexity. All these comparison algorithms are implemented in the same development environment, and utilize the Function Evaluations (FEs) as the stopping criterion. It is obvious that the more global optima the algorithm finds and the accuracy of these optima are higher when satisfying the stopping criterion, the better the algorithm performs.

\subsection{Test functions}

We use 20 multimodal benchmark functions to test these algorithms. Basic information of these test functions is summarized in Table \ref{tab:Test functions}, in which the symbol ``$-$'' in the last column corresponding to F16-F20 means that these functions have many local optima, and the number of their local optima are unknown. In Table \ref{tab:Test functions}, the former 15 multimodal functions come from CEC2015 \cite{Suganthan2015cec}, and the latter 5 functions are the classical multimodal functions with high dimension. These CEC2015 functions can be divided into two categories. The first 8 functions are expanded scalable functions and the remaining 7 functions are composition functions. All these CEC2015 functions come with search space shift and rotation that makes them more difficult to solve, while the latter 5 multimodal functions are only shifted. More details of these test functions are presented in the document named ``Definitions of CEC2015 niching benchmark 20141228'' which can be downloaded from the website shown in reference \cite{Suganthan2015cec}. For functions F2, F3, F5, F6, F7, F8, F9, F11, F12 and F13 the objective is to locate all the global optima, while for the rest the target is to escape from the local optima to hunt for the global optimum. And all these test functions are minimization problems.

\begin{table}[!hbp] \scriptsize
    \setlength{\abovecaptionskip}{0pt}
    \setlength{\belowcaptionskip}{10pt}
    \caption{Test functions}\label{tab:Test functions}
    \centering
    \begin{tabular}{c c c c c}
        \toprule[1px]
        Fn. & Test function name & Dimensions & No. of global optima & No. of local optima \\ \hline
        F1 & Expanded Two-Peak Trap & 5 & 1 & 15 \\
        F2 & Expanded Five-Uneven-Peak Trap & 5 & 32 & 0 \\
        F3 & Expanded Equal Minima & 4 & 625 & 0 \\
        F4 & Expanded Decreasing Minima & 5 & 1 & 15 \\
        F5 & Expanded Uneven Minima & 3 & 125 & 0 \\
        F6 & Expanded Himmelblau's Function & 4 & 16 & 0 \\
        F7 & Expanded Six-Hump Camel Back & 6 & 8 & 0 \\
        F8 & Modified Vincent Function & 3 & 216 & 0 \\
        F9 & Composition Function 1 & 10 & 10 & 0 \\
        F10 & Composition Function 2 & 10 & 1 & 9 \\
        F11 & Composition Function 3 & 10 & 10 & 0 \\
        F12 & Composition Function 4 & 10 & 10 & 0 \\
        F13 & Composition Function 5 & 10 & 10 & 0 \\
        F14 & Composition Function 6 & 10 & 1 & 19 \\
        F15 & Composition Function 7 & 10 & 1 & 19 \\
        F16 & Griewank & 50 & 1 & $-$ \\
        F17 & Ackley & 100 & 1 & $-$ \\
        F18 & Rosenbrock & 100 & 1 & $-$ \\
        F19 & Rastrigin & 100 & 1 & $-$ \\
        F20 & Expanded Scaffer's F6 & 100 & 1 & $-$ \\ \hline
        \multicolumn{5}{c}{Search range: ${[-100, 100]}^D$} \\
        \bottomrule[1px]
    \end{tabular}
\end{table}

\subsection{Parameters setting}

To compare the performance of the multimodal optimization algorithms in this paper, all the test functions should be treated as black-box problems, though their global optima can be obtained by the method of derivation. Thus, the known global optima of these test functions cannot be used by these algorithms during the iterations. The fitness error $\varepsilon_f$, i.e., level of accuracy, is used to judge whether the final solution is a real global optimum. If the difference between the fitness value of the final solution and the fitness value of the known global optimum is lower than $\varepsilon_f$, this solution can be considered a real global optimum. In our experiments, the fitness error $\varepsilon_f$, population size $p$ and function evaluations used by these algorithms for the test functions are listed in Table \ref{tab:Setting of parameters associated with test functions}. It is worth noting that a function which has higher dimension or more complex fitness landscape may require a larger population size or more function evaluations.

\begin{table}[!hbp]
    \setlength{\abovecaptionskip}{0pt}
    \setlength{\belowcaptionskip}{10pt}
    \caption{Setting of parameters associated with test functions}\label{tab:Setting of parameters associated with test functions}
    \centering
    \begin{tabular}{c c c c}
        \toprule[1px]
        Fn. & $\varepsilon_f$ & pop. size ($p$) & FEs \\ \hline
        F1 & 0.00000001 & 50 & 6.0E6 \\
        F2 & 0.00000001 & 50 & 1.8E8 \\
        F3 & 0.00000001 & 50 & 1.5E9 \\
        F4 & 0.00000001 & 50 & 1.5E8 \\
        F5 & 0.00000001 & 50 & 9.0E7 \\
        F6 & 0.00000001 & 50 & 3.0E7 \\
        F7 & 0.000001 & 50 & 3.0E7 \\
        F8 & 0.0001 & 50 & 1.5E9 \\
        F9 & 0.00000001 & 500 & 1.2E8 \\
        F10 & 0.00000001 & 500 & 3.0E7 \\
        F11 & 0.00000001 & 100 & 6.0E7 \\
        F12 & 0.00000001 & 100 & 5.0E7 \\
        F13 & 0.00000001 & 100 & 1.0E7 \\
        F14 & 0.00000001 & 500 & 5.0E7 \\
        F15 & 0.00000001 & 100 & 2.0E7 \\
        F16 & 0.00000001 & 100 & 2.0E7 \\
        F17 & 0.00000001 & 100 & 2.0E7 \\
        F18 & 0.00000001 & 100 & 1.5E8 \\
        F19 & 0.00000001 & 100 & 1.5E8 \\
        F20 & 0.00000001 & 100 & 6.0E7 \\
        \bottomrule[1px]
    \end{tabular}
\end{table}

The parameters' values of these comparison algorithms are set as same as those in their reference source respectively. Table \ref{tab:Setting of main parameters of algorithms} lists the values of main parameters of these algorithms.

\begin{table}[!hbp]
\centering
\begin{threeparttable}
    \setlength{\abovecaptionskip}{0pt}
    \setlength{\belowcaptionskip}{10pt}
    \caption{Setting of main parameters of algorithms}\label{tab:Setting of main parameters of algorithms}
    \begin{tabular*}{0.82\textwidth}{l l}
        \toprule[1px]
        Algorithms & Parameters \\ \hline
        LIPS & $\omega$=0.729844, $nsize$=2$~$5 \\
        NGSA & $G_\mathrm{0}$=10, $\alpha$=20, $k_i$=0.08, $k_f$=0.16 \\
        NSDE & $CR$=0.9, $F$=0.5 \\
        NCDE & $CR$=0.9, $F$=0.5 \\
        FERPSO & $\chi$=0.729844, $\varphi_\mathrm{max}$=4.1 \\
        WSA & $\rho _0$=2 \\
        WSA-IC & $\rho _0$=2, $\eta$=0, $T_s$=100$*n$, $T_f$=$\varepsilon_f$ \\
        \bottomrule[1px]
    \end{tabular*}
    \begin{tablenotes}
        \item[1.] $\omega$\text{: inertia weight;} $nsize$\text{: neighborhood size;}
        \item[2.] $G_\mathrm{0}$\text{: gravitational constant at the beginning;} $\alpha$\text{: attenuation coefficient;} $k_i$, $k_f$\text{: two constants that determine the number of neighbors at the beginning and at the end;}
        \item[3.] $CR$\text{: crossover rate;} $F$\text{: scaling factor;}
        \item[4.] $\chi$\text{: constriction factor;} $\varphi_\mathrm{max}$\text{: coefficient;}
    \end{tablenotes}
\end{threeparttable}
\end{table}

The attenuation coefficient $\eta$ of WSA for each test function is listed in Table \ref{tab:Attenuation coefficient of WSA for test functions}. Table 5 shows the neighborhood size $m$ of NSDE and NCDE respectively.

\begin{table}[!hbp] \footnotesize
    \setlength{\abovecaptionskip}{0pt}
    \setlength{\belowcaptionskip}{0pt}
    \renewcommand\arraystretch{1}
    \caption{Setting of attenuation coefficient of WSA for test functions}\label{tab:Attenuation coefficient of WSA for test functions}
    \centering
    \begin{tabular*}{\textwidth}{@{\extracolsep{\fill}}@{}c@{}c@{}c@{}c@{}c@{}c@{}c@{}c@{}c@{}c@{}c@{}c}
        \toprule[1px]
        Fn. & F1 & F2 & F3 & F4 & F5 & F6 & F7 & F8 & F9 & F10 \\ \hline
        $\eta$ & 0.0001 & 0.1 & 0.14 & 0.00005 & 0.16 & 0.16 & 0.001 & 0.3 & 0.09 & 0.001 \\ \hline
        Fn. & F11 & F12 & F13 & F14 & F15 & F16 & F17 & F18 & F19 & F20 \\ \hline
        $\eta$ & 0.01 & 0.001 & 0.001 & 0.001 & 0.001 & 0.005 & 0.01 & 0.014 & 0.005 & 0.01 \\
        \bottomrule[1px]
    \end{tabular*}
\end{table}

\begin{table}[!hbp] \scriptsize
    \setlength{\abovecaptionskip}{0pt}
    \setlength{\belowcaptionskip}{0pt}
    \renewcommand\arraystretch{1}
    \caption{Setting of neighborhood size $m$ of NSDE and NCDE for test functions}\label{tab:neighborhood size}
    \centering
    \begin{tabular*}{\textwidth}{@{\extracolsep{\fill}}@{}c@{}c@{}c@{}c@{}c@{}c@{}c@{}c@{}c@{}c@{}c@{}c@{}c@{}c@{}c@{}c@{}c@{}c@{}c@{}c@{}c}
        \toprule[1px]
        Fn. & F1 & F2 & F3 & F4 & F5 & F6 & F7 & F8 & F9 & F10 & F11 & F12 & F13 & F14 & F15 & F16 & F17 & F18 & F19 & F20 \\ \hline
        NSDE & 0.2$p$ & 0.2$p$ & 0.2$p$ & 0.2$p$ & 0.2$p$ & 0.2$p$ & 0.2$p$ & 0.2$p$ & 0.1$p$ & 0.1$p$ & 0.1$p$ & 0.1$p$ & 0.1$p$ & 0.1$p$ & 0.1$p$ & 0.1$p$ & 0.1$p$ & 0.1$p$ & 0.1$p$ & 0.1$p$  \\
        NCDE & 0.2$p$ & 0.2$p$ & 0.2$p$ & 0.2$p$ & 0.2$p$ & 0.2$p$ & 0.2$p$ & 0.2$p$ & 0.1$p$ & 0.1$p$ & 0.1$p$ & 0.1$p$ & 0.1$p$ & 0.1$p$ & 0.1$p$ & 0.1$p$ & 0.1$p$ & 0.1$p$ & 0.1$p$ & 0.1$p$ \\
        \bottomrule[1px]
    \end{tabular*}
\end{table}

\subsection{Performance metrics}

To fairly compare the performance of WSA-IC with that of other six algorithms, we have conducted 51 independent runs for each algorithm over each test function. And the following four metrics are used to measure the performance of all the algorithms.

\begin{enumerate}

\item[1)] Success Rate (SR) \cite{li2004adaptively}: the percentage of runs in which all the global optima are successfully located using the given level of accuracy.
\item[2)] Average Number of Optima Found (ANOF) \cite{Suganthan2015cec}: the average number of global optima found over 51 runs.
\item[3)] Quality of optima found: the mean of fitness values of optima found over 51 runs, reflecting the accuracy of optima found.
\item[4)] Convergence rate: the rate of an algorithm converging to the global optimum over function evaluations.

\end{enumerate}

\subsection{Quantity of optima found}

This section presents and analyses the results of quantity of optima found by these algorithms. Firstly, all the algorithms are compared on ``Success Rate'', which is the most popular metric used to test the performance of the multimodal optimization algorithms in terms of locating multiple global optima. Then, the metric ``Average Number of Optima Found'' is employed to further compare the performance of the algorithms on locating multiple global optima, as some algorithms can not achieve nonzero SR over some functions with multiple global optima.

\begin{enumerate}

\item[1)] Success Rate

\end{enumerate}

The SR of each algorithm on each test function is presented in Table \ref{tab:SR and ranks}, in which each number within the parenthese denotes the rank of each algorithm on the corresponding function in terms of SR, and the bold number means the corresponding algorithm performs best on the function. The same SR value on a function means that the corresponding algorithms have the same rank for the function. The last row of Table \ref{tab:SR and ranks} shows the total rank of each algorithm for all the test functions, which is the summation of each individual rank of the algorithm for each function. It can be seen from Table \ref{tab:SR and ranks} that WSA-IC performs best on most of the test functions in terms of SR. Especially on F3, F5 and F8 which have massive global optima, WSA-IC achieves the maximal SR values, i.e., 1, while the comparison algorithms can not achieve nonzero SR values on the three functions,indicating WSA-IC performs much better than other algorithms. It is worth noting that F9$-$F15 are composition functions with search space shift and rotation, whose global optima are more difficult to locate, so that all the algorithms can not achieve nonzero SR values on F9$-$F14. For the composition function F15, WSA-IC, LIPS, NSDE and NCDE all get the maximal SR value. What's more, for the high dimensional multimodal functions F16, F18 and F19, WSA-IC can also achieve much higher SR values than most of other multimodal optimization algorithms. It also can be seen that the better performance of WSA-IC in terms of SR can be supported by the total rank of WSA-IC which is much better than those achieved by other algorithms.

\begin{table}[htbp]
    \setlength{\abovecaptionskip}{0pt}
    \setlength{\belowcaptionskip}{0pt}
    \renewcommand\arraystretch{0.6}
    \caption{SR and ranks (in parentheses) of algorithms on F1$-$F20}\label{tab:SR and ranks}
    \centering
    \begin{tabular*}{\textwidth}{@{\extracolsep{\fill}}@{}c@{}c@{}c@{}c@{}c@{}c@{}c@{}c}
        \toprule[1px]
        Fn. & LIPS & NGSA & NSDE & NCDE & FERPSO & WSA & WSA-IC \\[2.5pt] \hline
        \multirow{2}{*}{F1} & 0.31 & 0.10 & 0.92 & 0.14 & 0.39 & 0.08 & \textbf{1} \\
        & (4) & (6) & (2) & (5) & (3) & (7) & \textbf{(1)} \\[2.5pt]
        \multirow{2}{*}{F2} & 0 & 0 & 0 & 0 & 0 & 0 & \textbf{1} \\
        & (2) & (2) & (2) & (2) & (2) & (2) & \textbf{(1)} \\[2.5pt]
        \multirow{2}{*}{F3} & 0 & 0 & 0 & 0 & 0 & 0 & \textbf{1} \\
        & (2) & (2) & (2) & (2) & (2) & (2) & \textbf{(1)} \\[2.5pt]
        \multirow{2}{*}{F4} & 0.31 & 0.49 & \textbf{(1)} & \textbf{(1)} & 0.14 & 0 & \textbf{1} \\
        & (5) & (4) & \textbf{(1)} & \textbf{(1)} & (6) & (7) & \textbf{(1)} \\[2.5pt]
        \multirow{2}{*}{F5} & 0 & 0 & 0 & 0 & 0 & 0 & \textbf{1} \\
        & (2) & (2) & (2) & (2) & (2) & (2) & \textbf{(1)} \\[2.5pt]
        \multirow{2}{*}{F6} & 0 & 0 & 0 & 0 & 0 & 0 & \textbf{1} \\
        & (2) & (2) & (2) & (2) & (2) & (2) & \textbf{(1)} \\[2.5pt]
        \multirow{2}{*}{F7} & 0 & 0 & 0 & 0.16 & 0 & 0 & \textbf{1} \\
        & (3) & (3) & (3) & (2) & (3) & (3) & \textbf{(1)} \\[2.5pt]
        \multirow{2}{*}{F8} & 0 & 0 & 0 & 0 & 0 & 0 & \textbf{1} \\
        & (2) & (2) & (2) & (2) & (2) & (2) & \textbf{(1)} \\[2.5pt]
        \multirow{2}{*}{F9} & \textbf{0} & \textbf{0} & \textbf{0} & \textbf{0} & \textbf{0} & \textbf{0} & \textbf{0} \\
        & \textbf{(1)} & \textbf{(1)} & \textbf{(1)} & \textbf{(1)} & \textbf{(1)} & \textbf{(1)} & \textbf{(1)} \\[2.5pt]
        \multirow{2}{*}{F10} & \textbf{0} & \textbf{0} & \textbf{0} & \textbf{0} & \textbf{0} & \textbf{0} & \textbf{0} \\
        & \textbf{(1)} & \textbf{(1)} & \textbf{(1)} & \textbf{(1)} & \textbf{(1)} & \textbf{(1)} & \textbf{(1)} \\[2.5pt]
        \multirow{2}{*}{F11} & \textbf{0} & \textbf{0} & \textbf{0} & \textbf{0} & \textbf{0} & \textbf{0} & \textbf{0} \\
        & \textbf{(1)} & \textbf{(1)} & \textbf{(1)} & \textbf{(1)} & \textbf{(1)} & \textbf{(1)} & \textbf{(1)} \\[2.5pt]
        \multirow{2}{*}{F12} & \textbf{0} & \textbf{0} & \textbf{0} & \textbf{0} & \textbf{0} & \textbf{0} & \textbf{0} \\
        & \textbf{(1)} & \textbf{(1)} & \textbf{(1)} & \textbf{(1)} & \textbf{(1)} & \textbf{(1)} & \textbf{(1)} \\[2.5pt]
        \multirow{2}{*}{F13} & \textbf{0} & \textbf{0} & \textbf{0} & \textbf{0} & \textbf{0} & \textbf{0} & \textbf{0} \\
        & \textbf{(1)} & \textbf{(1)} & \textbf{(1)} & \textbf{(1)} & \textbf{(1)} & \textbf{(1)} & \textbf{(1)} \\[2.5pt]
        \multirow{2}{*}{F14} & \textbf{0} & \textbf{0} & \textbf{0} & \textbf{0} & \textbf{0} & \textbf{0} & \textbf{0} \\
        & \textbf{(1)} & \textbf{(1)} & \textbf{(1)} & \textbf{(1)} & \textbf{(1)} & \textbf{(1)} & \textbf{(1)} \\[2.5pt]
        \multirow{2}{*}{F15} & \textbf{1} & 0.20 & \textbf{1} & \textbf{1} & 0.73 & 0 & \textbf{1} \\
        & \textbf{(1)} & (6) & \textbf{(1)} & \textbf{(1)} & (5) & (7) & \textbf{(1)} \\[2.5pt]
        \multirow{2}{*}{F16} & \textbf{1} & 0.12 & \textbf{1} & \textbf{1} & 0.39 & 0.41 & 0.98 \\
        & \textbf{(1)} & (7) & \textbf{(1)} & \textbf{(1)} & (6) & (5) & (4) \\[2.5pt]
        \multirow{2}{*}{F17} & \textbf{0} & 0 & \textbf{0.08} & \textbf{0} & \textbf{0} & \textbf{0} & \textbf{0} \\
        & (2) & (2) & \textbf{(1)} & (2) & (2) & (2) & (2) \\[2.5pt]
        \multirow{2}{*}{F18} & 0 & 0 & 0.82 & 0.88 & 0.24 & 0.57 & \textbf{0.88} \\
        & (6) & (6) & (1) & \textbf{(1)} & (5) & (4) & \textbf{(1)} \\[2.5pt]
        \multirow{2}{*}{F19} & 0 & 0 & 1 & 0 & 0 & 0 & \textbf{0.98} \\
        & (3) & (3) & \textbf{(1)} & (3) & (3) & (3) & (2) \\[2.5pt]
        \multirow{2}{*}{F20} & \textbf{0} & \textbf{0} & \textbf{0} & \textbf{0} & \textbf{0} & \textbf{0} & \textbf{0} \\
        & \textbf{(1)} & \textbf{(1)} & \textbf{(1)} & \textbf{(1)} & \textbf{(1)} & \textbf{(1)} & \textbf{(1)} \\[2.5pt] \hline
        Total rank & 42 & 54 & 30 & 33 & 50 & 55 & 25 \\
        \bottomrule[1px]
    \end{tabular*}
\end{table}

\begin{enumerate}

\item[2)] Average Number of Optima Found

\end{enumerate}

As the sample size in this paper is 51 that is greater than 30, we have conducted the Two Independent-samples Z-test for WSA-IC to judge whether the difference between its population and the population of every other algorithm, respectively represented by their independent samples, is significant or not on each test function under the significance level 0.05, which is based on the variance between the ANOF of two independent samples. Table \ref{tab:ANOF} presents the ANOF of each algorithm on each test function, and the standard deviation of the number of optima found is also listed. The symbol ``+'' means that the difference between the population of WSA-IC and the population of the comparison algorithm is significant, and WSA-IC performs better than the comparison algorithm, while the symbol ``='' means that the difference is not significant. And the symbol ``$-$'' means that the difference is significant, and WSA-IC performs worse than the comparison algorithm. The bold number in Table \ref{tab:ANOF} means that the corresponding algorithm performs best on the function in terms of ANOF. It can be seen from Table \ref{tab:ANOF} that WSA-IC has the best performance in terms of ANOF over F1$-$F8, F15 and F18, which echoes the best SR values of WSA-IC on these test functions as shown in Table 6. For the two composition functions F10 and F14 and the high dimensional function F20, all the algorithms can not get nonzero ANOF, which means that all the algorithms can not find the global optima of these functions. For the composition functions F9, F11 and F12, WSA-IC performs far better than most of other comparison algorithms in terms of ANOF. It also can be seen that the better performance of WSA-IC in terms of the number of optima found can be supported by the total number of symbols ``+'', ``='' and ``$-$'' in the last three rows of Table \ref{tab:ANOF}. As we can see from Table \ref{tab:ANOF}, the nonzero values of the number of symbol ``$-$'' only occur once when WSA-IC is compared with LIPS. And the number of symbol ``+'' is larger than that of symbol ``='' when compared with the other algorithms. The better performance of WSA-IC is firstly due to the improvement on the location update rule of WSA when $\eta$=0, i.e., a whale will move to a new position under the guidance of its ``better and nearest'' whale if this new position is better than its original position, which can ensure the formation of multiple subpopulations and maintain the ability of local exploitation. More importantly, the method of identifying and jumping out of the located extreme points during the iterations can improve the global search ability as much as possible, which can contribute significantly to the location of multiple global optima.

\begin{table}[htbp] \scriptsize
    \setlength{\abovecaptionskip}{0pt}
    \setlength{\belowcaptionskip}{0pt}
    \renewcommand\arraystretch{1}
    \caption{ANOF of algorithms on F1$-$F20}\label{tab:ANOF}
    \centering
    \begin{tabular*}{\textwidth}{@{\extracolsep{\fill}}|@{}c|@{}c|@{}c|@{}c|@{}c|@{}c|@{}c|@{}c|@{}c|@{}c|@{}c|@{}c|@{}c|@{}c|}
        \hline
        Fn. & \multicolumn{2}{c|}{LIPS} & \multicolumn{2}{c|}{NGSA} & \multicolumn{2}{c|}{NSDE} & \multicolumn{2}{c|}{NCDE} & \multicolumn{2}{c|}{FERPSO} & \multicolumn{2}{c|}{WSA} & WSA-IC \\[2.5pt] \hline
        F1 & 0.31$\pm$0.46 & + & 0.10$\pm$0.30 & + & 0.92$\pm$0.27 & = & 0.14$\pm$0.34 & + & 0.39$\pm$0.49 & + & 0.08$\pm$0.27 & + & \textbf{1$\pm$0} \\ \hline
        F2 & 10.86$\pm$1.36 & + & 2.84$\pm$1.04 & + & 1.51$\pm$0.50 & + & 0$\pm$0 & + & 2.67$\pm$0.88 & + & 0.76$\pm$0.47 & + & \textbf{32$\pm$0} \\ \hline
        F3 & 16.76$\pm$1.45 & + & 3.37$\pm$1.27 & + & 1.84$\pm$0.36 & + & 44.90$\pm$1.61 & + & 5.59$\pm$1.16 & + & 1.04$\pm$0.59 & + & \textbf{625$\pm$0} \\ \hline
        F4 & 0.31$\pm$0.46 & + & 0.49$\pm$0.50 & + & \textbf{1$\pm$0} & = & \textbf{1$\pm$0} & = & 0.14$\pm$0.34 & + & 0$\pm$0 & + & \textbf{1$\pm$0} \\ \hline
        F5 & 16.80$\pm$1.68 & + & 4.73$\pm$1.50 & + & 1.98$\pm$0.14 & + & 0.22$\pm$1.53 & + & 8.61$\pm$1.50 & + & 1.27$\pm$0.45 & + & \textbf{125$\pm$0} \\ \hline
        F6 & 9.65$\pm$1.49 & + & 4.37$\pm$0.93 & + & 2$\pm$0 & + & 7.96$\pm$1.83 & + & 4.25$\pm$1.10 & + & 0.92$\pm$0.39 & + & \textbf{16$\pm$0} \\ \hline
        F7 & 3.80$\pm$1.31 & + & 1.27$\pm$0.89 & + & 2$\pm$0 & + & 5.90$\pm$1.47 & + & 1.49$\pm$0.70 & + & 0.47$\pm$0.50 & + & \textbf{8$\pm$0} \\ \hline
        F8 & 16.04$\pm$1.67 & + & 8.75$\pm$2.09 & + & 2.02$\pm$0.14 & + & 33.24$\pm$4.04 & + & 7.90$\pm$1.47 & + & 0.69$\pm$0.98 & + & \textbf{216$\pm$0} \\ \hline
        F9 & \textbf{6.31$\pm$0.67} & $-$ & 0.61$\pm$0.56 & + & 0$\pm$0 & + & 2$\pm$0 & + & 0.82$\pm$0.68 & + & 1.51$\pm$0.54 & + & 4.53$\pm$1.04 \\ \hline
        F10 & \textbf{0$\pm$0} & = & \textbf{0$\pm$0} & = & \textbf{0$\pm$0} & = & \textbf{0$\pm$0} & = & \textbf{0$\pm$0} & = & \textbf{0$\pm$0} & = & \textbf{0$\pm$0} \\ \hline
        F11 & 0.51$\pm$0.50 & + & 0.51$\pm$0.50 & + & 0.02$\pm$0.14 & + & \textbf{0.84$\pm$0.36} & = & 0.02$\pm$0.14 & + & 0.33$\pm$0.47 & + & 0.82$\pm$0.38 \\ \hline
        F12 & \textbf{0.18$\pm$0.38} & = & 0$\pm$0 & = & 0$\pm$0 & = & 0$\pm$0 & = & 0$\pm$0 & = & 0$\pm$0 & = & 0.04$\pm$0.19 \\ \hline
        F13 & 0.96$\pm$0.19 & = & 0.06$\pm$0.24 & + & \textbf{1$\pm$0} & = & \textbf{1$\pm$0} & = & 0.76$\pm$0.42 & = & 0$\pm$0 & + & 0.90$\pm$0.30 \\ \hline
        F14 & \textbf{0$\pm$0} & = & \textbf{0$\pm$0} & = & \textbf{0$\pm$0} & = & \textbf{0$\pm$0} & = & \textbf{0$\pm$0} & = & \textbf{0$\pm$0} & = & \textbf{0$\pm$0} \\ \hline
        F15 & \textbf{1$\pm$0} & = & 0.20$\pm$0.40 & + & \textbf{1$\pm$0} & = & \textbf{1$\pm$0} & = & 0.73$\pm$0.45 & + & 0$\pm$0 & + & \textbf{1$\pm$0} \\ \hline
        F16 & \textbf{1$\pm$0} & = & 0.12$\pm$0.32 & + & \textbf{1$\pm$0} & = & \textbf{1$\pm$0} & = & 0.39$\pm$0.49 & + & 0.41$\pm$0.49 & + & 0.98$\pm$0.14 \\ \hline
        F17 & \textbf{0$\pm$0} & = & 0$\pm$0 & = & 0.08$\pm$0.27 & = & \textbf{0$\pm$0} & = & \textbf{0$\pm$0} & = & \textbf{0$\pm$0} & = & \textbf{0$\pm$0} \\ \hline
        F18 & 0$\pm$0 & + & 0$\pm$0 & + & 0.82$\pm$0.38 & + & \textbf{0.88$\pm$0.32} & + & 0.24$\pm$0.42 & + & 0.57$\pm$0.50 & + & \textbf{0.88$\pm$0.32} \\ \hline
        F19 & 0$\pm$0 & + & 0$\pm$0 & + & \textbf{1$\pm$0} & + & 0$\pm$0 & + & 0$\pm$0 & + & 0$\pm$0 & + & 0.98$\pm$0.14 \\ \hline
        F20 & \textbf{0$\pm$0} & = & \textbf{0$\pm$0} & = & \textbf{0$\pm$0} & = & \textbf{0$\pm$0} & = & \textbf{0$\pm$0} & = & \textbf{0$\pm$0} & = & \textbf{0$\pm$0} \\ \hline
        + & \multicolumn{2}{c|}{11} & \multicolumn{2}{c|}{15} & \multicolumn{2}{c|}{8} & \multicolumn{2}{c|}{9} & \multicolumn{2}{c|}{14} & \multicolumn{2}{c|}{15} & \\ \hline
        = & \multicolumn{2}{c|}{8} & \multicolumn{2}{c|}{5} & \multicolumn{2}{c|}{12} & \multicolumn{2}{c|}{11} & \multicolumn{2}{c|}{6} & \multicolumn{2}{c|}{5} & \\ \hline
        $-$ & \multicolumn{2}{c|}{1} & \multicolumn{2}{c|}{0} & \multicolumn{2}{c|}{0} & \multicolumn{2}{c|}{0} & \multicolumn{2}{c|}{1} & \multicolumn{2}{c|}{0} & \\ \hline
    \end{tabular*}
\end{table}

\subsection{Quality of optima found}

This section compares the performance of these algorithms in terms of the quality of optima found. Table \ref{tab:Quality of optima found} presents the mean of fitness values of optima found over 51 runs on all these test functions, and the standard deviation of fitness values of optima found are also listed in the parentheses. For comparing the performance of all the algorithms on the quality of optima found, 100*Fn. (Fn. denotes the serial number of a function) is substracted from the fitness values of optima found by all the algorithms on the CEC2015 niching test functions (i.e., F1$-$F15 in Table \ref{tab:Test functions}). And we have also conducted the Two Independent-samples Z-test between WSA-IC and other comparison algorithms. The bold number in Table \ref{tab:Quality of optima found} means that the corresponding algorithm performs best on the function in terms of the quality of optima found. It can be seen form Table \ref{tab:Quality of optima found} that WSA-IC has the best performance over F1, F4, F7 and F14. What's more, WSA-IC shows very stable performanc in terms of the quality of optima found over these functions, which can be supported by the total number of symbols ``+'', ``='' and ``$-$'' in the last three rows of Table \ref{tab:Quality of optima found}, in which the number of symbol ``$-$'' corresponding to different comparison algorithms is much less than that of symbols ``+'' and ``=''.

\begin{table}[htbp] \tiny
    \setlength{\abovecaptionskip}{0pt}
    \setlength{\belowcaptionskip}{0pt}
    \renewcommand\arraystretch{1}
    \caption{Quality of optima found by algorithms on F1$-$F20}\label{tab:Quality of optima found}
    \centering
    \begin{tabular*}{\textwidth}{@{\extracolsep{\fill}}|@{}c|@{}c|@{}c|@{}c|@{}c|@{}c|@{}c|@{}c|@{}c|@{}c|@{}c|@{}c|@{}c|@{}c|}
        \hline
        Fn. & \multicolumn{2}{c|}{LIPS} & \multicolumn{2}{c|}{NGSA} & \multicolumn{2}{c|}{NSDE} & \multicolumn{2}{c|}{NCDE} & \multicolumn{2}{c|}{FERPSO} & \multicolumn{2}{c|}{WSA} & WSA-IC \\[2.5pt] \hline
        \multirow{2}{*}{F1} & 2.65E+01 & \multirow{2}{*}{+} & 5.18E+01 & \multirow{2}{*}{+} & 3.14E+00 & \multirow{2}{*}{+} & 6.53E+00 & \multirow{2}{*}{+} & 2.98E+01 & \multirow{2}{*}{+} & 8.71E+01 & \multirow{2}{*}{+} & \textbf{0.00E+00} \\
        & (2.01E+01) & & (2.66E+01) & & (1.08E+01) & & (1.37E+01) & & (2.73E+01) & & (4.39E+01) & & \textbf{(0.00E+00)} \\ \hline
        \multirow{2}{*}{F2} & \textbf{0.00E+00} & \multirow{2}{*}{=} & 1.58E$-$10 & \multirow{2}{*}{=} & 1.17E$-$10 & \multirow{2}{*}{=} & 6.86E$-$03 & \multirow{2}{*}{=} & 2.26E$-$13 & \multirow{2}{*}{=} & 9.32E+00 & \multirow{2}{*}{+} & 4.88E$-$16 \\
        & \textbf{(0.00E+00)} & & (3.81E$-$10) & & (5.07E$-$10) & & (1.52E$-$02) & & (7.85E$-$13) & & (1.97E+01) & & (2.11E$-$15) \\ \hline
        \multirow{2}{*}{F3} & \textbf{0.00E+00} & \multirow{2}{*}{=} & 4.66E$-$09 & \multirow{2}{*}{=} & 3.90E$-$15 & \multirow{2}{*}{=} & 1.60E$-$11 & \multirow{2}{*}{=} & 8.41E$-$13 & \multirow{2}{*}{=} & 1.57E$-$01 & \multirow{2}{*}{=} & 3.42E$-$16 \\
        & \textbf{(0.00E+00)} & & (3.14E$-$08) & & (9.78E$-$15) & & (5.53E$-$11) & & (3.44E$-$12) & & (3.64E$-$01) & & (5.28E$-$16) \\ \hline
        \multirow{2}{*}{F4} & 7.86E$-$02 & \multirow{2}{*}{+} & 6.17E$-$02 & \multirow{2}{*}{=} & \textbf{0.00E+00} & \multirow{2}{*}{=} & 1.89E$-$14 & \multirow{2}{*}{=} & 1.71E$-$01 & \multirow{2}{*}{+} & 1.52E+00 & \multirow{2}{*}{+} & \textbf{0.00E+00} \\
        & (6.68E$-$02) & & (7.11E$-$02) & & \textbf{(0.00E+00)} & & (3.31E$-$14) & & (1.33E$-$01) & & (5.64E$-$01) & & \textbf{(0.00E+00)} \\ \hline
        \multirow{2}{*}{F5} & \textbf{0.00E+00} & \multirow{2}{*}{=} & 2.72E¨C10 & \multirow{2}{*}{=} & 5.57E$-$16 & \multirow{2}{*}{=} & 3.88E$-$06 & \multirow{2}{*}{=} & 3.74E$-$13 & \multirow{2}{*}{=} & 6.69E$-$15 & \multirow{2}{*}{=} & 1.52E$-$16 \\
        & \textbf{(0.00E+00)} & & (4.20E$-$10) & & (3.94E$-$15) & & (3.38E$-$06) & & (8.30E$-$13) & & (1.83E$-$14) & & (3.89E$-$16) \\ \hline
        \multirow{2}{*}{F6} & \textbf{0.00E+00} & \multirow{2}{*}{=} & 1.91E$-$10 & \multirow{2}{*}{=} & 1.11E$-$15 & \multirow{2}{*}{=} & 1.53E$-$14 & \multirow{2}{*}{=} & 1.19E$-$13 & \multirow{2}{*}{=} & 5.84E$-$01 & \multirow{2}{*}{+} & 2.37E$-$15 \\
        & \textbf{(0.00E+00)} & & (5.55E$-$10) & & (7.88E$-$15) & & (3.21E$-$14) & & (1.72E$-$13) & & (2.42E+00) & & (5.39E$-$15) \\ \hline
        \multirow{2}{*}{F7} & \textbf{5.58E$-$07} & \multirow{2}{*}{=} & 7.04E$-$01 & \multirow{2}{*}{+} & \textbf{5.58E$-$07} & \multirow{2}{*}{=} & 5.65E$-$07 & \multirow{2}{*}{=} & 6.40E$-$02 & \multirow{2}{*}{=} & 2.41E+00 & \multirow{2}{*}{+} & \textbf{5.58E$-$07} \\
        & \textbf{(8.76E$-$15)} & & (1.34E+00) & & \textbf{(0.00E+00)} & & (2.75E$-$08) & & (4.53E$-$01) & & (2.95E+00) & & \textbf{(0.00E+00)} \\ \hline
        \multirow{2}{*}{F8} & \textbf{0.00E+00} & \multirow{2}{*}{=} & 6.26E$-$06 & \multirow{2}{*}{=} & 7.73E$-$09 & \multirow{2}{*}{=} & 1.05E$-$05 & \multirow{2}{*}{=} & 2.02E$-$11 & \multirow{2}{*}{=} & 5.38E$-$01 & \multirow{2}{*}{+} & 2.09E$-$08 \\
        & \textbf{(0.00E+00)} & & (6.35E$-$06) & & (5.47E$-$08) & & (6.30E$-$06) & & (1.27E$-$10) & & (1.13E+00) & & (6.67E$-$08) \\ \hline
        \multirow{2}{*}{F9} & 1.78E$-$13 & \multirow{2}{*}{=} & 5.50E$-$01 & \multirow{2}{*}{+} & 1.53E+00 & \multirow{2}{*}{+} & \textbf{0.00E+00} & \multirow{2}{*}{=} & 4.81E$-$01 & \multirow{2}{*}{+} & 2.02E$-$10 & \multirow{2}{*}{=} & 3.77E$-$14 \\
        & (1.15E$-$13) & & (7.45E$-$01) & & (0.00E+00) & & \textbf{(0.00E+00)} & & (7.12E$-$01) & & (5.69E$-$10) & & (2.92E$-$14) \\ \hline
        \multirow{2}{*}{F10} & \textbf{3.00E+01} & \multirow{2}{*}{$-$} & 1.39E+04 & \multirow{2}{*}{+} & \textbf{3.00E+01} & \multirow{2}{*}{$-$} & \textbf{3.00E+01} & \multirow{2}{*}{$-$} & 9.84E+03 & \multirow{2}{*}{+} & 1.07E+04 & \multirow{2}{*}{+} & 3.82E+01 \\
        & \textbf{(2.22E$-$12)} & & (1.10E$-$11) & & \textbf{(0.00E+00)} & & \textbf{(5.24E$-$05)} & & (6.32E+03) & & (5.89E+03) & & (1.34E+01) \\ \hline
        \multirow{2}{*}{F11} & 4.61E$-$03 & \multirow{2}{*}{=} & 8.26E$-$02 & \multirow{2}{*}{=} & 2.04E$-$01 & \multirow{2}{*}{+} & \textbf{3.00E$-$03} & \multirow{2}{*}{=} & 3.78E$-$01 & \multirow{2}{*}{+} & 3.44E$-$01 & \multirow{2}{*}{+} & 3.78E$-$02 \\
        & (1.44E$-$02) & & (2.80E$-$01) & & (1.07E$-$01) & & \textbf{(1.02E$-$02)} & & (3.52E$-$01) & & (7.03E$-$01) & & (8.59E$-$02) \\ \hline
        \multirow{2}{*}{F12} & 5.61E+01 & \multirow{2}{*}{+} & 7.04E+02 & \multirow{2}{*}{+} & \textbf{4.69E$-$02} & \multirow{2}{*}{$-$} & 8.83E$-$02 & \multirow{2}{*}{$-$} & 3.27E+01 & \multirow{2}{*}{+} & 3.88E+02 & \multirow{2}{*}{+} & 4.41E+00 \\
        & (1.47E+02) & & (2.25E+02) & & \textbf{(1.86E$-$02)} & & (1.07E$-$01) & & (8.17E+01) & & (2.71E+02) & & (2.47E+01) \\ \hline
        \multirow{2}{*}{F13} & 9.58E+00 & \multirow{2}{*}{$-$} & 2.89E+02 & \multirow{2}{*}{+} & 4.15E$-$13 & \multirow{2}{*}{$-$} & \textbf{0.00E+00} & \multirow{2}{*}{$-$} & 6.01E+01 & \multirow{2}{*}{+} & 3.91E+02 & \multirow{2}{*}{+} & 2.28E+01 \\
        & (4.74E+01) & & (8.44E+01) & & (1.47E$-$13) & & \textbf{(0.00E+00)} & & (1.12E+02) & & (8.24E+01) & & (6.98E+01) \\ \hline
        \multirow{2}{*}{F14} & 2.12E+02 & \multirow{2}{*}{+} & 4.22E+02 & \multirow{2}{*}{+} & 1.12E+02 & \multirow{2}{*}{+} & 8.02E+01 & \multirow{2}{*}{+} & 2.13E+02 & \multirow{2}{*}{+} & 6.20E+02 & \multirow{2}{*}{+} & \textbf{5.91E+01} \\
        & (2.15E+02) & & (4.51E+01) & & (7.36E+01) & & (7.09E+00) & & (1.41E+02) & & (1.10E+02) & & \textbf{(4.16E+01)} \\ \hline
        \multirow{2}{*}{F15} & 2.54E$-$13 & \multirow{2}{*}{=} & 1.73E+02 & \multirow{2}{*}{+} & 4.99E$-$13 & \multirow{2}{*}{=} & \textbf{0.00E+00} & \multirow{2}{*}{=} & 5.92E+01 & \multirow{2}{*}{+} & 2.91E+02 & \multirow{2}{*}{+} & 6.24E$-$14 \\
        & (1.32E$-$13) & & (9.93E+01) & & (1.86E$-$13) & & \textbf{(0.00E+00)} & & (9.71E+01) & & (4.95E+01) & & (2.92E$-$13) \\ \hline
        \multirow{2}{*}{F16} & \textbf{7.58E$-$14} & \multirow{2}{*}{=} & 3.59E$-$01 & \multirow{2}{*}{+} & 2.63E$-$13 & \multirow{2}{*}{=} & 1.74E$-$13 & \multirow{2}{*}{=} & 1.28E$-$01 & \multirow{2}{*}{=} & 1.11E$-$02 & \multirow{2}{*}{=} & 4.83E$-$04 \\
        & \textbf{(1.07E$-$13)} & & (3.34E$-$01) & & (8.27E$-$14) & & (9.64E$-$14) & & (2.82E$-$01) & & (1.23E$-$02) & & (1.95E$-$03) \\ \hline
        \multirow{2}{*}{F17} & 2.11E+01 & \multirow{2}{*}{+} & 2.13E+01 & \multirow{2}{*}{+} & \textbf{2.43E$-$03} & \multirow{2}{*}{$-$} & 2.11E+01 & \multirow{2}{*}{+} & 2.00E+01 & \multirow{2}{*}{=} & 2.00E+01 & \multirow{2}{*}{=} & 2.00E+01 \\
        & (3.18E$-$02) & & (2.56E$-$02) & & \textbf{(8.54E$-$03)} & & (6.69E$-$02) & & (1.78E$-$02) & & (2.91E$-$04) & & (3.41E$-$03) \\ \hline
        \multirow{2}{*}{F18} & 1.52E+00 & \multirow{2}{*}{$-$} & 2.39E+08 & \multirow{2}{*}{+} & 6.44E$-$01 & \multirow{2}{*}{$-$} & \textbf{4.69E$-$01} & \multirow{2}{*}{$-$} & 7.37E+07 & \multirow{2}{*}{+} & 1.72E+00 & \multirow{2}{*}{$-$} & 1.77E+02 \\
        & (4.05E+00) & & (1.98E+08) & & (2.42E+00) & & \textbf{(1.28E+00)} & & (9.80E+07) & & (1.97E+00) & & (1.22E+03) \\ \hline
        \multirow{2}{*}{F19} & 1.28E+02 & \multirow{2}{*}{+} & 4.00E+03 & \multirow{2}{*}{+} & \textbf{1.01E$-$12} & \multirow{2}{*}{$-$} & 3.65E+02 & \multirow{2}{*}{+} & 2.55E+03 & \multirow{2}{*}{+} & 5.69E+03 & \multirow{2}{*}{+} & 1.50E+00 \\
        & (2.06E+01) & & (1.79E+03) & & \textbf{(1.45E$-$13)} & & (1.58E+02) & & (2.61E+03) & & (1.44E+03) & & (1.06E+01) \\ \hline
        \multirow{2}{*}{F20} & 3.05E+01 & \multirow{2}{*}{$-$} & 4.63E+01 & \multirow{2}{*}{+} & \textbf{1.83E+00} & \multirow{2}{*}{$-$} & 4.43E+01 & \multirow{2}{*}{+} & 3.72E+01 & \multirow{2}{*}{$-$} & 4.44E+01 & \multirow{2}{*}{+} & 4.36E+01 \\
        & (2.38E+00) & & (3.68E$-$01) & & \textbf{(1.02E+00)} & & (1.75E+00) & & (1.66E+00) & & (7.13E$-$01) & & (7.79E$-$01) \\ \hline
        + & \multicolumn{2}{c|}{6} & \multicolumn{2}{c|}{13} & \multicolumn{2}{c|}{4} & \multicolumn{2}{c|}{5} & \multicolumn{2}{c|}{11} & \multicolumn{2}{c|}{14} & \\ \hline
        = & \multicolumn{2}{c|}{10} & \multicolumn{2}{c|}{7} & \multicolumn{2}{c|}{9} & \multicolumn{2}{c|}{11} & \multicolumn{2}{c|}{8} & \multicolumn{2}{c|}{5} & \\ \hline
        $-$ & \multicolumn{2}{c|}{4} & \multicolumn{2}{c|}{0} & \multicolumn{2}{c|}{7} & \multicolumn{2}{c|}{4} & \multicolumn{2}{c|}{1} & \multicolumn{2}{c|}{1} & \\ \hline
    \end{tabular*}
\end{table}

\begin{figure}[htbp]
    \setlength{\abovecaptionskip}{0pt}
    \includegraphics[width=1\textwidth]{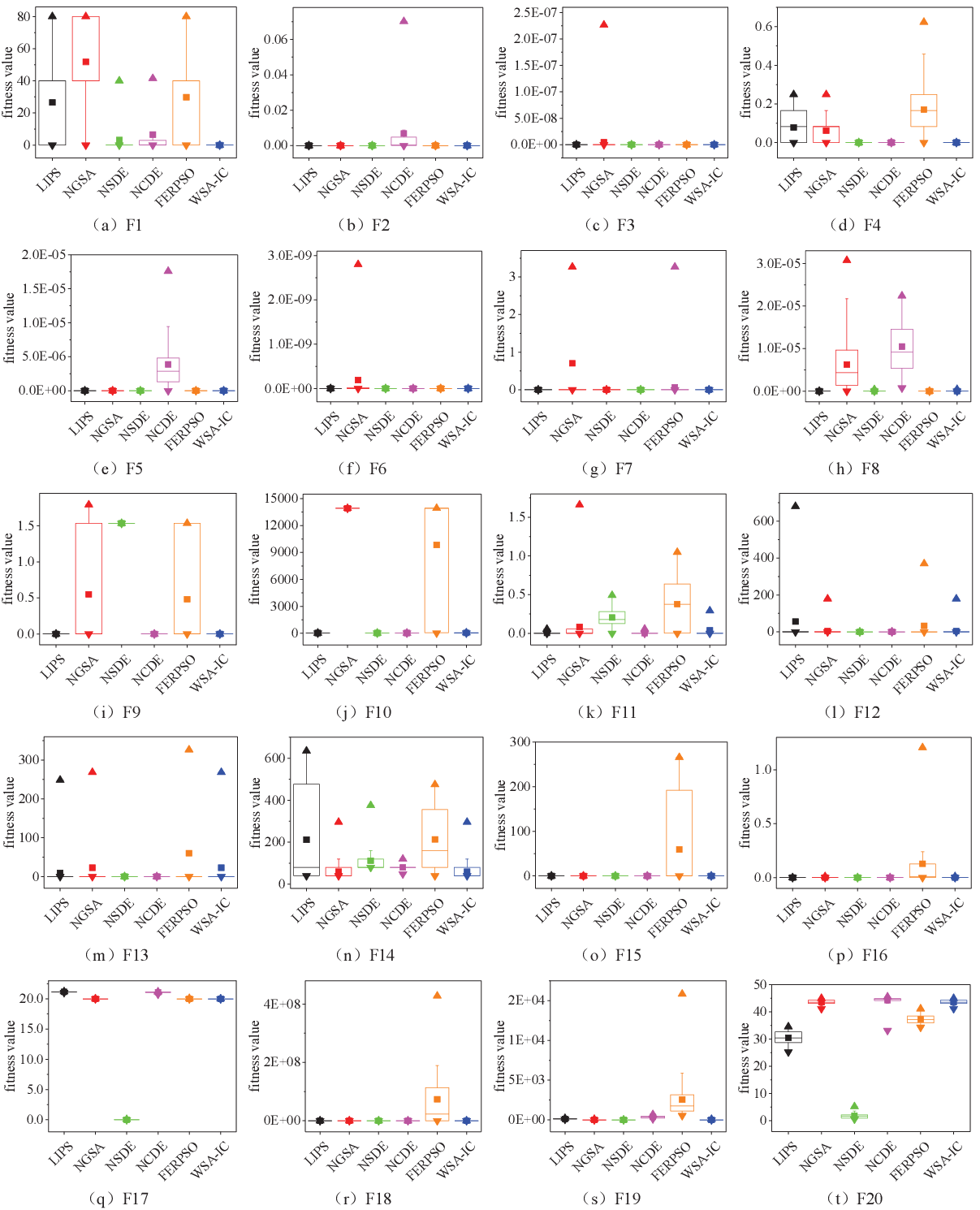}
    \caption{Box plot of algorithms on F1$-$F20}
    \label{fig:Box plot}
\end{figure}

What's more, the box plot of mean fitness values of optima found per run over 51 runs, by WSA-IC, LIPS, NGSA, NSDE, NCDE and FERPSO, is shown in Fig. \ref{fig:Box plot}. Since the quality of optima found by WSA are worse than other algorithms over most of these functions as shown in table 8, the blox plot of WSA are ignored, so as to ensure the obvious differences of other algorithms in terms of the distribution of optima found. It can be seen from Fig. \ref{fig:Box plot} that, the dispersion degree of mean fitness values of optima found by WSA-IC is quite small on most of the test functions with respect to other comparison algorithms. And WSA-IC only has outliers on F11, F12, F13, F14 and F20, while most of other algorithms have more outliers over these test functions. Therefore, it can be concluded that WSA-IC has good stability on the  accuracy of optima found over these test functions, with respect to other comparison algorithms. The better performance of WSA-IC in terms of the quality of optima found is also due to the improvement on the location update rule of WSA, i.e., a whale moves to a new position under the guidance of its ``better and nearest'' whale if this new position is better than its original position, which can ensure the ability of local exploitation. For example, when some whales follow the same extreme point, the best whale among these whales will stay where it is with great probability to guide other whales to converge to the extreme point followed by them. Besides, the method of identifying and jumping out of the located extreme points during the iterations can improve the global search ability as much as possible to find the global optima. For example, if some whales converge to a solution that is close to a global optimum, with this method some other whales that have reached steady state will be reinitialized, and they may move to the positions that is close to those convergent whales, which will accelerate these whales to converge to the global optimum.

\subsection{Convergence rate}

\begin{figure}[htbp]
    \setlength{\abovecaptionskip}{5pt}
    \subfigure[F1]{
        \label{fig:Convergence:a}
        \begin{minipage}[c]{0.32\textwidth}
            \centering
            \includegraphics[width=1.0\textwidth]{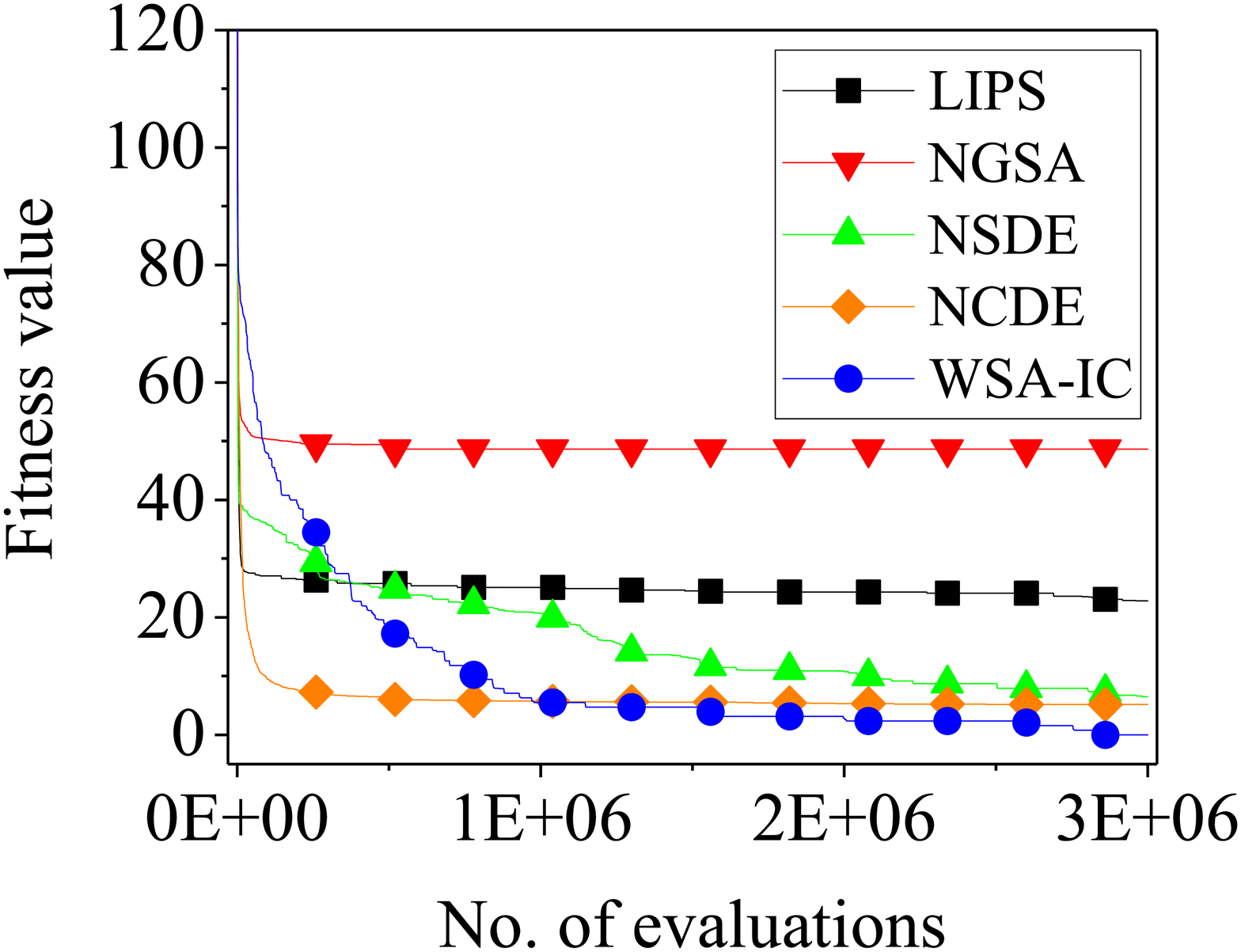}
        \end{minipage}}%
    \subfigure[F4]{
        \label{fig:Convergence:b}
        \begin{minipage}[c]{0.32\textwidth}
            \centering
            \includegraphics[width=1.0\textwidth]{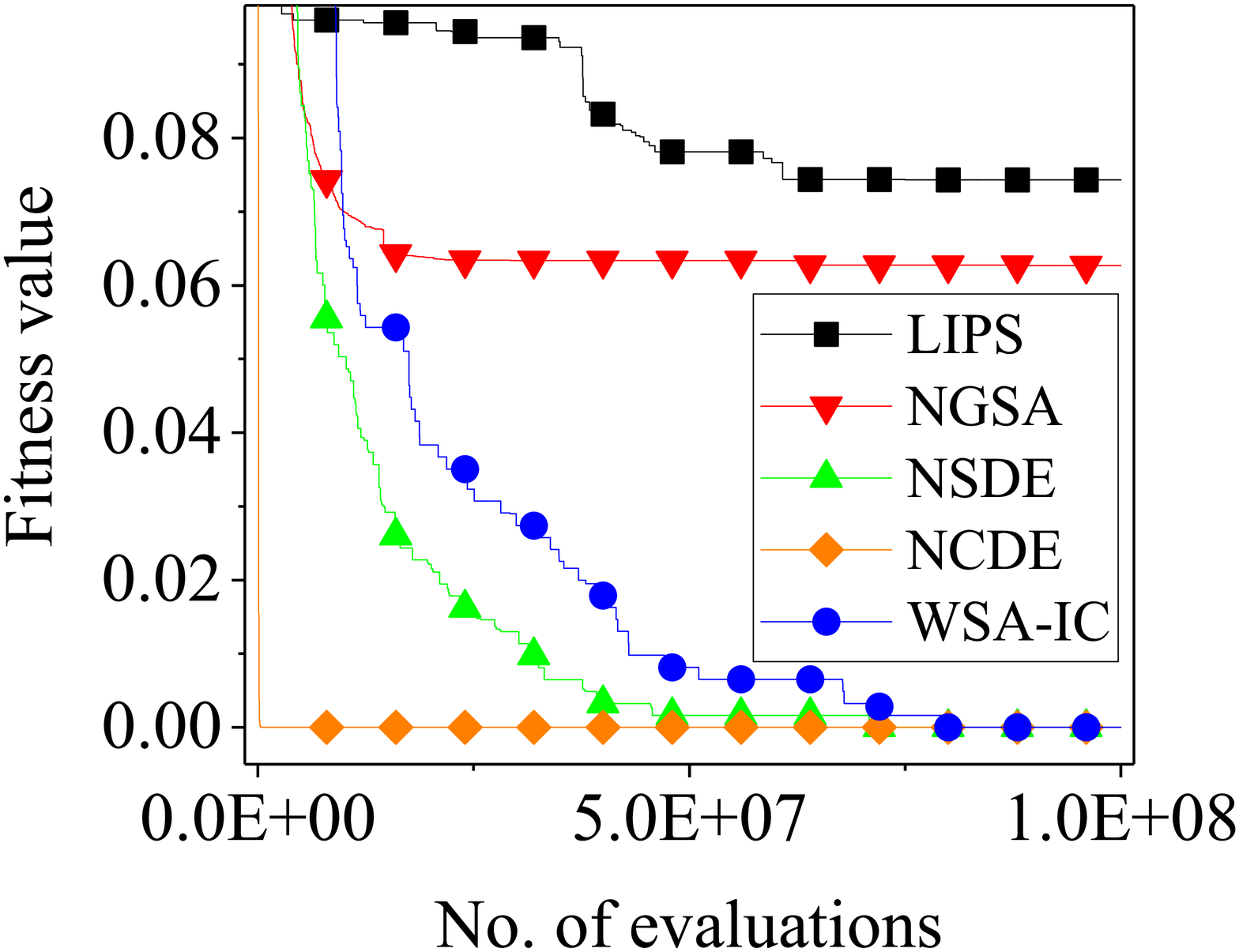}
        \end{minipage}}
    \subfigure[F9]{
        \label{fig:Convergence:c}
        \begin{minipage}[c]{0.32\textwidth}
            \centering
            \includegraphics[width=1.0\textwidth]{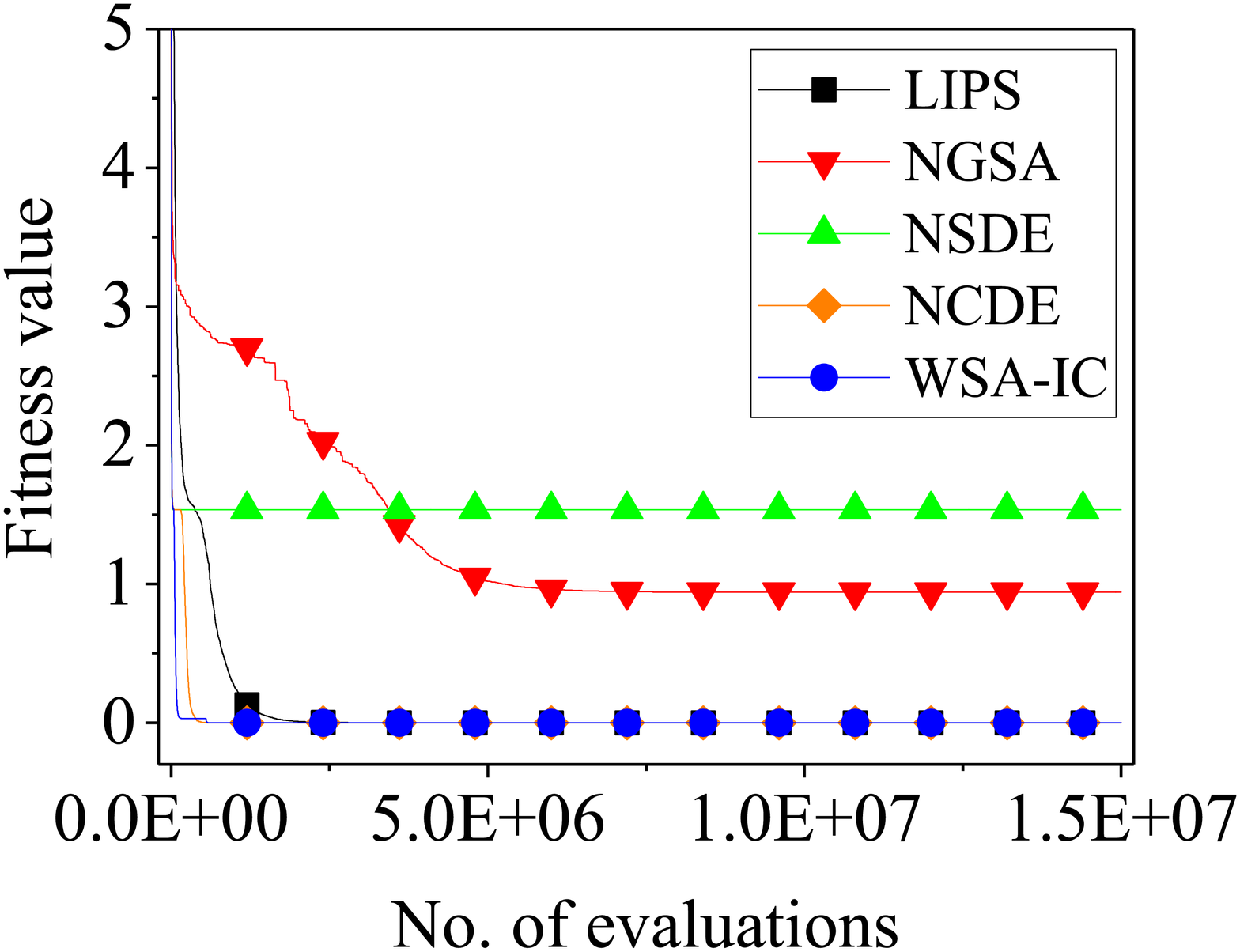}
        \end{minipage}}\\
    \subfigure[F14]{
        \label{fig:Convergence:d}
        \begin{minipage}[c]{0.32\textwidth}
            \centering
            \includegraphics[width=1.0\textwidth]{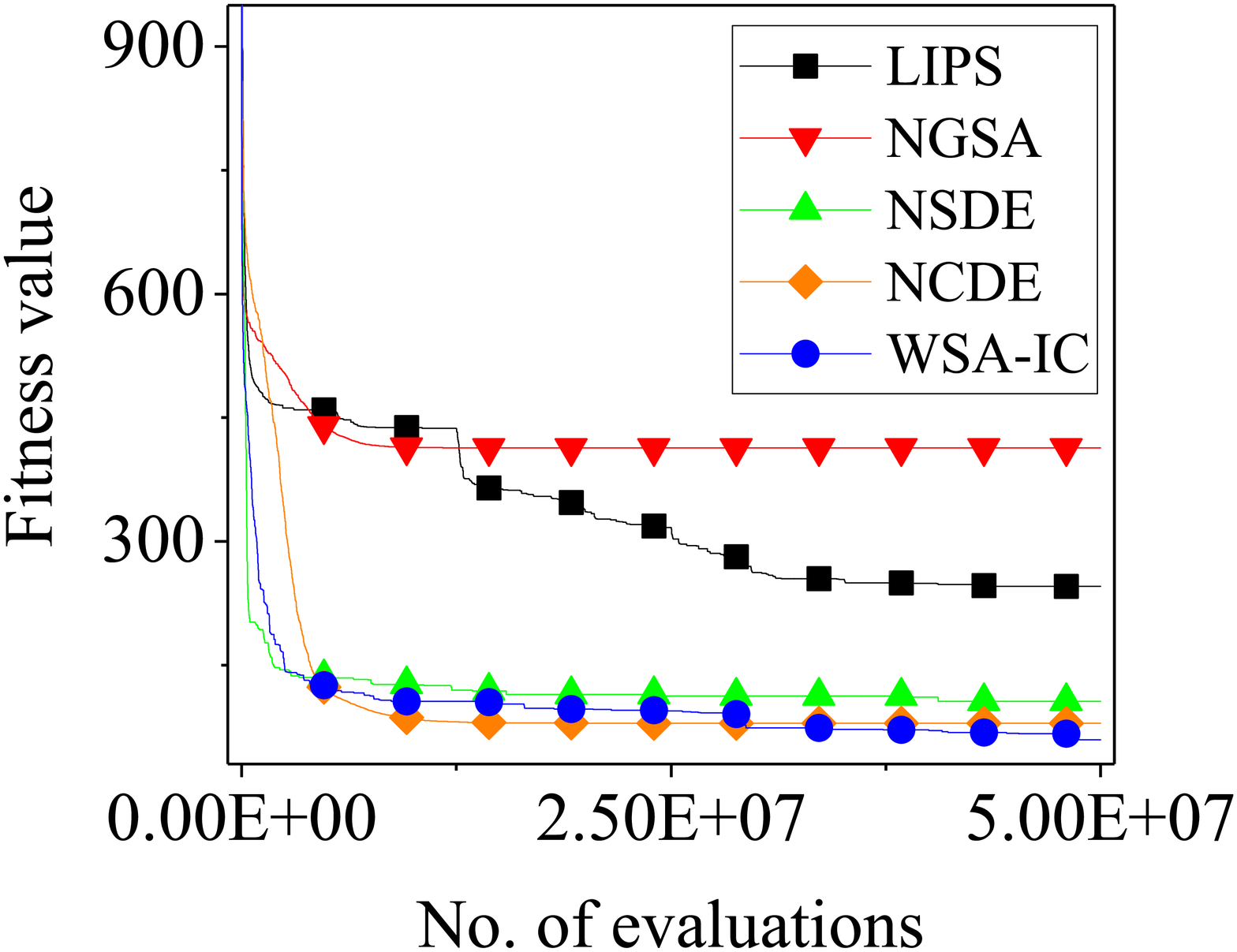}
        \end{minipage}}%
    \subfigure[F18]{
        \label{fig:Convergence:e}
        \begin{minipage}[c]{0.32\textwidth}
            \centering
            \includegraphics[width=1.0\textwidth]{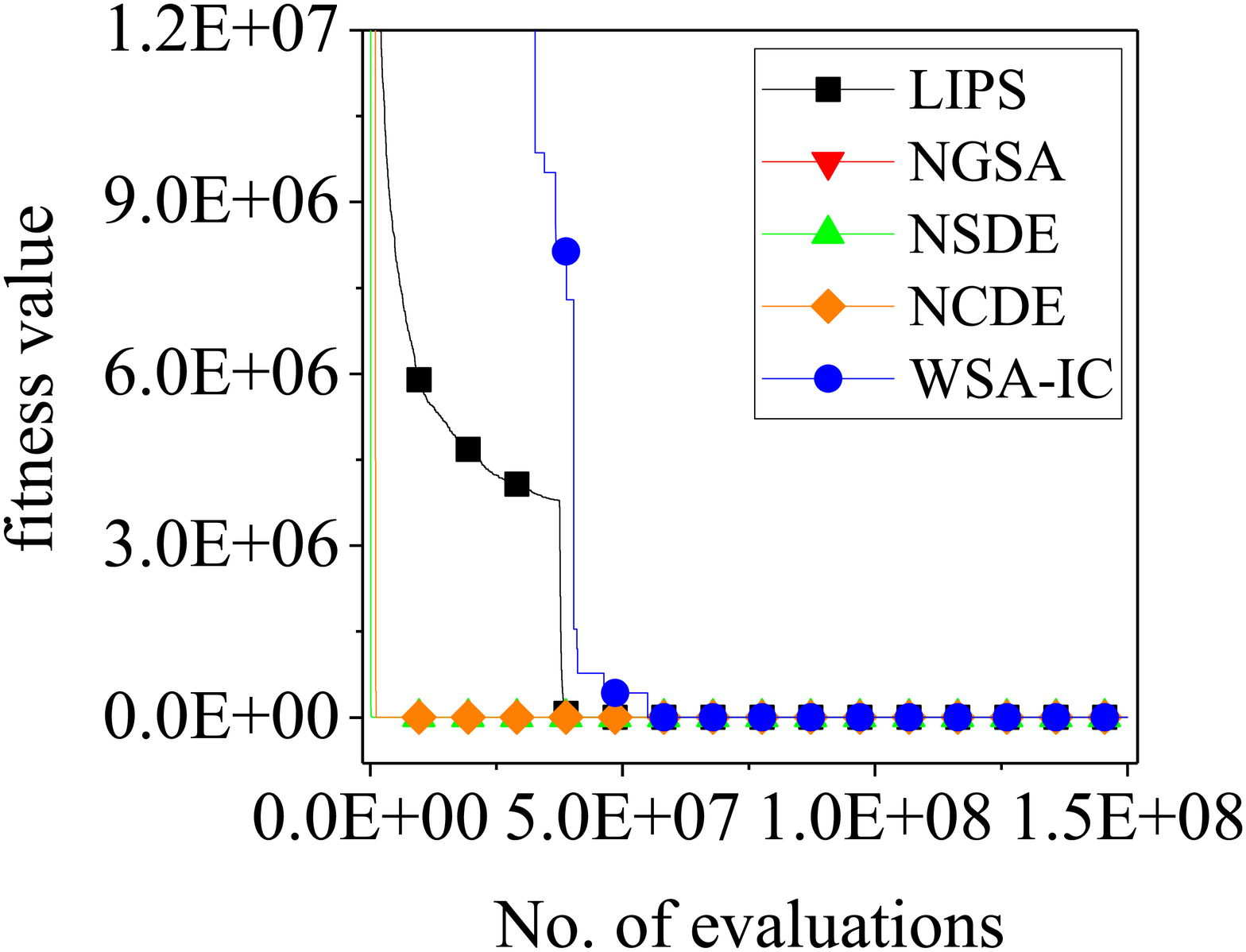}
        \end{minipage}}
    \subfigure[F19]{
        \label{fig:Convergence:f}
        \begin{minipage}[c]{0.32\textwidth}
            \centering
            \includegraphics[width=1.0\textwidth]{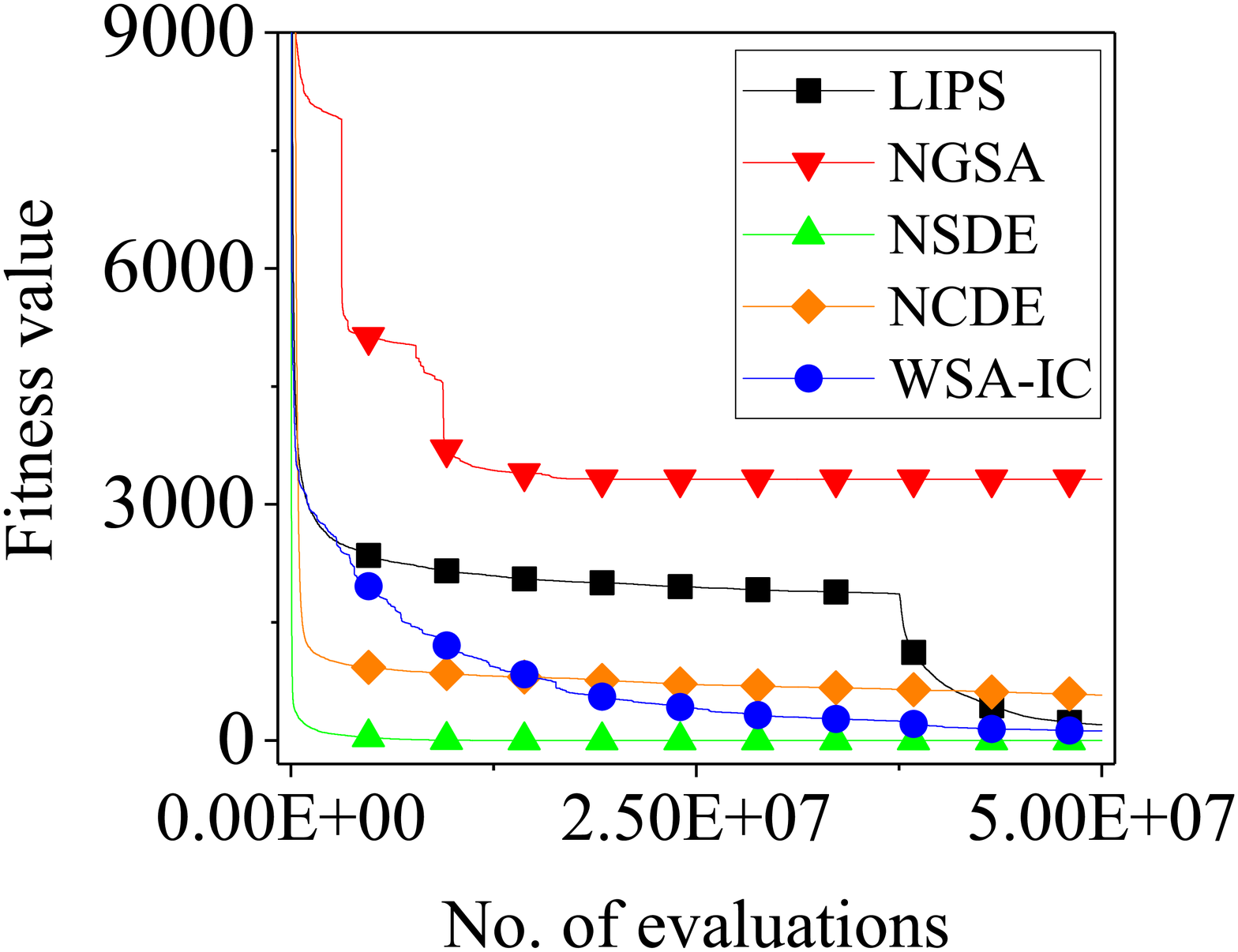}
        \end{minipage}}
    \caption{Convergence rate of algorithms on F1, F4, F9, F14, F18 and F19}
    \label{fig:Convergence}
\end{figure}

From the previous two sections, it can be seen that the proposed WSA-IC has better and more consistent performance than other algorithms, in terms of both the quantity of optima found and the quality of optima found on most test functions. To demonstrate the efficiency of WSA-IC on locating the global optima, WSA-IC is compared with other algorithms except FERPSO and WSA (because the population of FERPSO and WSA may prematurely converge to a solution or several solutions with same fitness value and terminate the iteration) in terms of convergence rate in this section. Six functions (i.e., F1, F4, F9, F14, F18 and F19, wherein F9 has no local optima while others all come with local optima) are used to test these algorithms. The convergence curves of all the algorithms on these test functions are depicted in Fig. \ref{fig:Convergence}, in which the horizontal axis represent the number of function evaluations and the vertical axis denote the mean of fitness values of the current global optima over 51 runs. It can be seen from Fig. \ref{fig:Convergence:c} that, for function F9 without local optima, NSDE cannot converge to the global optima, and WSA-IC converge to the global optima with much faster rate than that of LIPS and NCDE. Although NCDE can converge to the global optima of F9, it gets a much lower ANOF on F9 than that gained by WSA-IC as shown in Table 7. What's more, for functions F1, F4, F14, F18 and F19 that have multiple local optima, WSA-IC can achieve the global optima with satisfying convergence rate on F4 and F18 as shown in Fig. \ref{fig:Convergence:b} and Fig. \ref{fig:Convergence:e}. For F19, WSA-IC only performs a little worse than NSDE and far better than other algorithms, as shown in Fig. \ref{fig:Convergence:f}. And WSA-IC can obtain better solutions with faster convergence rate than other algorithms on F1 and F14, as shown in Fig. \ref{fig:Convergence:a} and Fig. \ref{fig:Convergence:d}. Therefore, it can be concluded that the proposed WSA-IC shows excellent performance on convergence rate relative to other algorithms.

\subsection{Discussion of WSA-IC parameters}

As mentioned in section \ref{sec:Parameters setting}, the parameters $\rho_0$ and $\eta$ are two constants, and are always set to 2 and 0 respectively. For almost all the problems, especially those problems without prior knowledge, $T_f$ can be set to ${\rm{1}}{\rm{.0}} \times {\rm{1}}{{\rm{0}}^{{\rm{ - 8}}}}$. Thus, only the parameter stability threshold $T_s$ may need to be specified different values for different problems. This section presents the results of ANOF obtained by WSA-IC on all these test functions with different $T_s$ values, as shown in Table \ref{tab:ANOF with different stability threshold}. And a clear visual comparison of ANOF obtained by WSA-IC with different $T_s$ values is shown in Fig. \ref{fig:Overview of ANOF}, where the values of ANOF with different $T_s$ values on each test function are normalized, and 1 refers to the best ANOF value while 0 refers to the worst ANOF value. It can be seen from Table \ref{tab:ANOF with different stability threshold} and Fig. \ref{fig:Overview of ANOF} that, WSA-IC can achieve the best ANOF values on most test functions with $T_s$=100$n$. Therefore, the parameter $T_s$ can be set to 100$n$ for almost all the continuous optimization problems.

\begin{table}[htbp] \footnotesize
    \setlength{\abovecaptionskip}{0pt}
    \setlength{\belowcaptionskip}{0pt}
    \renewcommand\arraystretch{1}
    \caption{ANOF of WSA-IC with different $T_s$ values on F1$-$F20}\label{tab:ANOF with different stability threshold}
    \centering
    \begin{tabular*}{\textwidth}{@{\extracolsep{\fill}}@{}c@{}c@{}c@{}c@{}c@{}c@{}c@{}c@{}c@{}c@{}c}
        \toprule[1px]
        Fn. & $T_s$=20$n$ & $T_s$=40$n$ & $T_s$=60$n$ & $T_s$=80$n$ & $T_s$=100$n$ & $T_s$=120$n$ & $T_s$=140$n$ & $T_s$=160$n$ & $T_s$=180$n$ & $T_s$=200$n$ \\ \hline
        F1 & \textbf{1} & \textbf{1} & \textbf{1} & \textbf{1} & \textbf{1} & \textbf{1} & \textbf{1} & \textbf{1} & \textbf{1} & \textbf{1} \\
        F2 & \textbf{32} & \textbf{32} & \textbf{32} & \textbf{32} & \textbf{32} & \textbf{32} & \textbf{32} & \textbf{32} & \textbf{32} & \textbf{32} \\
        F3 & 477.22 & 588.67 & 622.35 & 624.96 & \textbf{625} & \textbf{625} & \textbf{625} & \textbf{625} & \textbf{625} & 624.98 \\
        F4 & 0.76 & 0.84 & 0.88 & \textbf{1} & \textbf{1} & \textbf{1} & \textbf{1} & \textbf{1} & \textbf{1} & \textbf{1} \\
        F5 & \textbf{125} & \textbf{125} & \textbf{125} & \textbf{125} & \textbf{125} & 124.98 & \textbf{125} & \textbf{125} & \textbf{125} & \textbf{125} \\
        F6 & \textbf{16} & \textbf{16} & \textbf{16} & \textbf{16} & \textbf{16} & \textbf{16} & \textbf{16} & \textbf{16} & \textbf{16} & \textbf{16} \\
        F7 & \textbf{8} & \textbf{8} & \textbf{8} & \textbf{8} & \textbf{8} & \textbf{8} & \textbf{8} & \textbf{8} & \textbf{8} & 7.98 \\
        F8 & 215.98 & 215.98 & 215.98 & \textbf{216} & \textbf{216} & 215.92 & 215.96 & 215.94 & 215.94 & 215.80 \\
        F9 & \textbf{5.86} & 4.94 & 4.51 & 4.12 & 4.53 & 4.10 & 4.43 & 4.20 & 4.06 & 4.04 \\
        F10 & \textbf{0} & \textbf{0} & \textbf{0} & \textbf{0} & \textbf{0} & \textbf{0} & \textbf{0} & \textbf{0} & \textbf{0} & \textbf{0} \\
        F11 & \textbf{0.82} & 0.80 & 0.80 & 0.76 & \textbf{0.82} & 0.75 & 0.67 & 0.67 & 0.63 & 0.67 \\
        F12 & 0.02 & 0.02 & \textbf{0.04} & 0.02 & \textbf{0.04} & 0 & 0.02 & 0 & 0 & 0 \\
        F13 & \textbf{1} & 0.98 & 0.90 & 0.82 & 0.90 & 0.86 & 0.82 & 0.76 & 0.82 & 0.69 \\
        F14 & \textbf{0} & \textbf{0} & \textbf{0} & \textbf{0} & \textbf{0} & \textbf{0} & \textbf{0} & \textbf{0} & \textbf{0} & \textbf{0} \\
        F15 & 0.98 & \textbf{1} & 0.73 & 0.94 & \textbf{1} & 0.84 & 0.90 & 0.76 & 0.82 & 0.82 \\
        F16 & 0.84 & 0.76 & 0.92 & 0.88 & \textbf{0.98} & 0.88 & 0.73 & 0.92 & 0.90 & 0.88 \\
        F17 & \textbf{0} & \textbf{0} & \textbf{0} & \textbf{0} & \textbf{0} & \textbf{0} & \textbf{0} & \textbf{0} & \textbf{0} & \textbf{0} \\
        F18 & 0.82 & 0.86 & 0.86 & 0.84 & \textbf{0.88} & 0.86 & 0.86 & 0.73 & 0.71 & 0.72 \\
        F19 & \textbf{1} & 0.98 & 0.98 & 0.98 & 0.98 & 0.92 & 0.82 & 0.88 & 0.84 & 0.92 \\
        F20 & \textbf{0} & \textbf{0} & \textbf{0} & \textbf{0} & \textbf{0} & \textbf{0} & \textbf{0} & \textbf{0} & \textbf{0} & \textbf{0} \\
        \bottomrule[1px]
    \end{tabular*}
\end{table}

\begin{figure}[htbp]
    \setlength{\abovecaptionskip}{0pt}
    \includegraphics[width=1\textwidth]{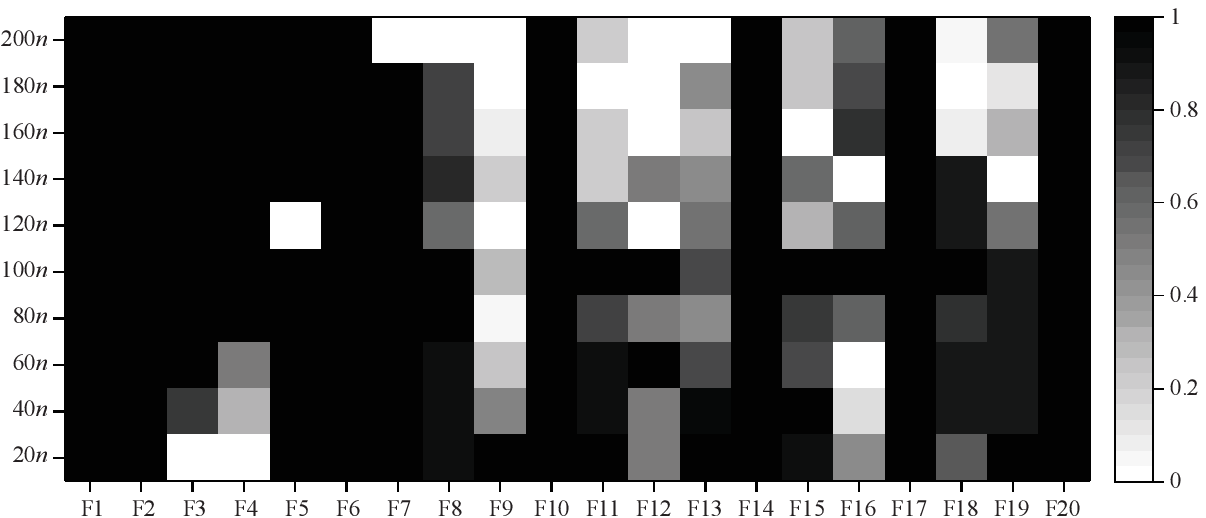}
    \caption{Overview of ANOF obtained by WSA-IC with different $T_s$ values on each function}
    \label{fig:Overview of ANOF}
\end{figure}

\section{Conclusions and future research}\label{sec:Conclusions}

A new multimodal optimizer named Whale Swarm Algorithm with Iterative Counter (WSA-IC), based on our preliminary work in \cite{Zeng2017}, is proposed in this paper. Firstly, WSA-IC improves the iteration rule of the original WSA when attenuation coefficient $\eta$ is set to 0, i.e., a whale moves to a new position under the guidance of its ``better and nearest'' whale if this new position is better than its original position. As a result, WSA-IC removes the need of specifying different values of $\eta$ for different problems to form multiple subpopulations, without introducing any niching parameters. And the ability of local exploitation is also ensured. What's more, WSA-IC enables the identification of extreme points and enables jumping out of the located extreme points during the iterations, relying on two new parameters, i.e., stability threshold $T_s$ and fitness threshold $T_f$. If a whale does not find a better position after successive $T_s$ iterations, it is considered to have located an extreme point and is to be reinitialized, so as to eliminate the unnecessary function evaluations and improve the global search ability. If the difference between the fitness value of the located extreme point and $f_{gbest}$ (the fitness value of the best one among the current global optima) is less than $T_f$, the located extreme point is considered a current global optimum. The values of $T_s$ and $T_f$ are very easy to set for different problems. Moreover, the convergence of WSA-IC is proved. The experimental results clearly show that WSA-IC performs statistically better than other niching metaheuristic algorithms over most test functions in terms of comprehensive metrics.

The main contributions of this paper are summarized into four aspects.

\begin{enumerate}

\item[1)] WSA-IC removes the need of specifying optimal niching parameter for different problems, which increases the practicality.
\item[2)] WSA-IC can efficiently identify and jump out of the located extreme points during the iterations, so as to locate as more global optima as possible in a single run, which further increases the practicality.
\item[3)] The algorithm dependent parameters of WSA-IC are easy to set for different problems, which also increases the practicality.
\item[4)] The population size of WSA-IC does not need to match the number of optima of the optimization problem. Generally, WSA-IC can keep a relative small population size, which contributes significantly to reducing the computation complexity.

\end{enumerate}

In the future, we will focus on the following aspects.

\begin{enumerate}

\item[1)] Introduce other metaheuristic algorithms or heuristic algorithms for the current best whale to execute the neighborhood search process in each iteration, so as to further improve the local search ability and the quality of optima.
\item[2)] Design some new methods to escape from the located extreme points instead of random reinitialization, to make the population spread over the entire solution space as much as possible.

\end{enumerate}

\paragraph{{\rm \textbf{Funding:}}}
This study was funded by the National Natural Science Foundation of China (NSFC) (51825502, 51775216 and 51721092), Natural Science Foundation of Hubei Province (2018CFA078) and the Program for HUST Academic Frontier Youth Team.

\paragraph{{\rm \textbf{Conflict of interest:}}}
We declare that we have no financial and personal relationships with other people or organizations that can inappropriately influence our work, there is no professional or other personal interest of any nature or kind in any product, service and/or company that could be construed as influencing the position presented in, or the review of, the manuscript entitled ``Whale swarm algorithm with the mechanism of identifying and escaping from extreme points for multimodal function optimization''.


\begin{thebibliography}{}

\bibitem{tasgetiren2017iterated}
Tasgetiren MF, Kizilay D, Pan Q-K, Suganthan PN (2017) Iterated greedy algorithms for the blocking flowshop scheduling problem with makespan criterion. Comput Oper Res 77:111-126

\bibitem{lin2016effective}
Lin G, Zhu W, Ali MM (2016) An effective hybrid memetic algorithm for the minimum weight dominating set problem. IEEE T Evolut Comput 20(6):892-907

\bibitem{zhang2017Object}
Zhang H, Cao X, Ho JKL, Chow TWS. (2017) Object-level video advertising: an optimization framework. IEEE T Ind Inform 13(2):520-531

\bibitem{ciancio2016heuristic}
Ciancio C, Ambrogio G, Gagliardi F, Musmanno R (2016) Heuristic techniques to optimize neural network architecture in manufacturing applications. Neural Comput Appl 27(7):2001-2015

\bibitem{csevkli2017multi}
{\c{S}}evkli AZ, G{\"u}ler B (2017) A multi-phase oscillated variable neighbourhood search algorithm for a real-world open vehicle routing problem. Appl Soft Comput 58:128-144

\bibitem{yi2016parallel}
Yi J, Li X, Chu C-H, Gao L (2016) Parallel chaotic local search enhanced harmony search algorithm for engineering design optimization. J Intell Manuf :1-24

\bibitem{palmieri2017comparison}
Raja MAZ, Ahmed U, Zameer A, Kiani AK, Chaudhary NI (2017) Bio-inspired heuristics hybrid with sequential quadratic programming and interior-point methods for reliable treatment of economic load dispatch problem. Neural Comput Appl. doi:10.1007/s00521-017-3019-3

\bibitem{Zhang2012Nature}
Zhang H, Llorca J, Davis CC, Milner SD (2012) Nature-inspired self-organization, control, and optimization in heterogeneous wireless networks. IEEE T Mobile Comput 11(7):1207-1222

\bibitem{li2010niching}
Li X (2010) Niching without niching parameters: particle swarm optimization using a ring topology. IEEE T Evolut Comput 14(1):150-169

\bibitem{de1975analysis}
De Jong KA (1975) Analysis of the behavior of a class of genetic adaptive systems

\bibitem{goldberg1987genetic}
Goldberg DE, Richardson J (1987) Genetic algorithms with sharing for multimodal function optimization. In: Genetic algorithms and their applications: Proceedings of the Second International Conference on Genetic Algorithms. Hillsdale, NJ: Lawrence Erlbaum, pp 41-49

\bibitem{yin1993fast}
Yin X, Germay N (1993) A fast genetic algorithm with sharing scheme using cluster analysis methods in multimodal function optimization. In: Artificial neural nets and genetic algorithms. Springer, pp 450-457

\bibitem{harik1995finding}
Harik GR (1995) Finding Multimodal Solutions Using Restricted Tournament Selection. In: ICGA. pp 24-31

\bibitem{bessaou2000island}
Bessaou M, P¨¦trowski A, Siarry P (2000) Island model cooperating with speciation for multimodal optimization. In: International Conference on Parallel Problem Solving from Nature. Springer, pp 437-446

\bibitem{deb1989investigation}
Deb K, Goldberg DE (1989) An investigation of niche and species formation in genetic function optimization. In: Proceedings of the 3rd international conference on genetic algorithms. Morgan Kaufmann Publishers Inc., pp 42-50

\bibitem{kennedy2002population}
Kennedy J, Mendes R (2002) Population structure and particle swarm performance. In: Evolutionary Computation, 2002. CEC'02. Proceedings of the 2002 Congress on. IEEE, pp 1671-1676

\bibitem{li2016seeking}
Li X, Epitropakis M, Deb K, Engelbrecht A (2016) Seeking multiple solutions: an updated survey on niching methods and their applications. IEEE T Evolut Comput 21(4): 518-538

\bibitem{thomsen2004multimodal}
Thomsen R (2004) Multimodal optimization using crowding-based differential evolution. In: Evolutionary Computation, 2004. CEC2004. Congress on. IEEE, pp 1382-1389

\bibitem{mahfoud1992crowding}
Mahfoud SW (1992) Crowding and preselection revisited. Urbana 51:61801

\bibitem{mengshoel1999probabilistic}
Mengshoel OJ, Goldberg DE (1999) Probabilistic crowding: Deterministic crowding with probabilistic replacement. In: Proc. of the Genetic and Evolutionary Computation Conference (GECCO-99). p 409

\bibitem{ursem1999multinational}
Ursem RK (1999) Multinational evolutionary algorithms. In: Evolutionary Computation, 1999. CEC 99. Proceedings of the 1999 Congress on. IEEE, pp 1633-1640

\bibitem{stoean2007disburdening}
Stoean CL, Preuss M, Stoean R, Dumitrescu D (2007) Disburdening the species conservation evolutionary algorithm of arguing with radii. In: Proceedings of the 9th annual conference on Genetic and evolutionary computation. ACM, pp 1420-1427

\bibitem{Zeng2017}
Zeng B, Gao L, Li X (2017) Whale Swarm Algorithm for Function Optimization. In: Huang D-S, Bevilacqua V, Premaratne P, Gupta P (eds) Intelligent Computing Theories and Application: 13th International Conference, ICIC 2017, Liverpool, UK, August 7-10, 2017, Proceedings, Part I. Springer International Publishing, Cham, pp 624-639. doi:10.1007/978-3-319-63309-1\_55

\bibitem{das2011real}
Das S, Maity S, Qu B-Y, Suganthan PN (2011) Real-parameter evolutionary multimodal optimization¡ªA survey of the state-of-the-art. Swarm Evol Compu 1(2):71-88

\bibitem{li2002species}
Li J-P, Balazs ME, Parks GT, Clarkson PJ (2002) A species conserving genetic algorithm for multimodal function optimization. Evol Comput 10(3):207-234

\bibitem{li2005efficient}
Li X (2005) Efficient differential evolution using speciation for multimodal function optimization. In: Proceedings of the 7th annual conference on Genetic and evolutionary computation. ACM, pp 873-880

\bibitem{li2004adaptively}
Li X (2004) Adaptively choosing neighbourhood bests using species in a particle swarm optimizer for multimodal function optimization. In: Genetic and Evolutionary Computation¨CGECCO 2004. Springer, pp 105-116

\bibitem{beasley1993sequential}
Beasley D, Bull DR, Martin RR (1993) A sequential niche technique for multimodal function optimization. Evol Comput 1(2):101-125

\bibitem{brits2002niching}
Brits R, Engelbrecht AP, Van den Bergh F (2002) A niching particle swarm optimizer. In: Proceedings of the 4th Asia-Pacific conference on simulated evolution and learning. Singapore: Orchid Country Club, pp 692-696

\bibitem{stoean2010multimodal}
Stoean C, Preuss M, Stoean R, Dumitrescu D (2010) Multimodal optimization by means of a topological species conservation algorithm. IEEE T Evolut Comput 14(6):842-864

\bibitem{deb2010finding}
Deb K, Saha A (2010) Finding multiple solutions for multimodal optimization problems using a multi-objective evolutionary approach. In: Proceedings of the 12th annual conference on genetic and evolutionary computation. ACM, pp 447-454

\bibitem{li2015history}
Li L, Tang K (2015) History-based topological speciation for multimodal optimization. IEEE T Evolut Comput 19(1):136-150

\bibitem{liang2006comprehensive}
Liang JJ, Qin AK, Suganthan PN, Baskar S (2006) Comprehensive learning particle swarm optimizer for global optimization of multimodal functions. IEEE T Evolut Comput 10(3):281-295

\bibitem{li2007multimodal}
Li X (2007) A multimodal particle swarm optimizer based on fitness Euclidean-distance ratio. In: Proceedings of the 9th annual conference on Genetic and evolutionary computation. ACM, pp 78-85

\bibitem{qu2012differential}
Qu B-Y, Suganthan PN, Liang J-J (2012) Differential evolution with neighborhood mutation for multimodal optimization. IEEE T Evolut Comput 16(5):601-614

\bibitem{qu2013distance}
Qu B-Y, Suganthan P, Das S (2013) A distance-based locally informed particle swarm model for multimodal optimization. IEEE T Evolut Comput 17(3):387-402

\bibitem{Yazdani2014gravitational}
Yazdani S, Nezamabadi-pour H, Kamyab S (2014) A gravitational search algorithm for multimodal optimization. Swarm Evol Comput 14:1-14

\bibitem{wang2015mommop}
Wang Y, Li H-X, Yen GG, Song W (2015) MOMMOP: Multiobjective optimization for locating multiple optimal solutions of multimodal optimization problems. IEEE T Cybernetics 45(4):830-843

\bibitem{Suganthan2015cec}
\url{http://www.ntu.edu.sg/home/EPNSugan/index_files/CEC2015/CEC2015.htm}

\end{thebibliography}
\end{document}